%% file: main.tex
\documentclass{article}

\usepackage{microtype}
\usepackage{graphicx}
\usepackage{graphicx, caption, subcaption}
\usepackage{booktabs} 

\usepackage [latin1]{inputenc}

\usepackage[accepted]{icml2024}

\usepackage[colorlinks,linkcolor=black,citecolor=blue, pagebackref=true]{hyperref}
\renewcommand*{\backrefalt}[4]{%
    \ifcase #1 \footnotesize{(Not cited.)}%
    \or        \footnotesize{(Cited on page~#2.)}%
    \else      \footnotesize{(Cited on pages~#2.)}%
    \fi}
\usepackage{bookmark}

\usepackage{amsmath}
\usepackage{amssymb}
\usepackage{mathtools}
\usepackage{amsthm}

\usepackage[capitalize,noabbrev]{cleveref}

\theoremstyle{plain}
\newtheorem{theorem}{Theorem}[section]

\newtheorem{lemma}[theorem]{Lemma}

\theoremstyle{definition}

\theoremstyle{remark}

\usepackage[textsize=tiny]{todonotes}

\input{preambleicml2024}
\icmltitlerunning{CompeteSMoE - Effective Training of Sparse Mixture of Experts via Competition}

\begin{document}

\twocolumn[
\icmltitle{CompeteSMoE - Effective Sparse Mixture of Experts Training via Competition}


\icmlsetsymbol{equal}{*}

\begin{icmlauthorlist}
\icmlauthor{Quang Pham}{smu}
\icmlauthor{Giang Do}{ten}
\icmlauthor{Huy Nguyen}{texas}
\icmlauthor{TrungTin Nguyen}{queen}
\icmlauthor{Chenghao Liu}{sf}
\icmlauthor{Mina Sartipi}{ten}
\icmlauthor{Binh T. Nguyen}{hcmus}
\icmlauthor{Savitha Ramasamy}{astar}
\icmlauthor{Xiaoli Li}{astar}
\icmlauthor{Steven Hoi}{equal,smu}
\icmlauthor{Nhat Ho}{equal,texas}
\end{icmlauthorlist}

\icmlaffiliation{smu}{Singapore Management University}
\icmlaffiliation{ten}{University of Tennessee at Chattanooga}
\icmlaffiliation{texas}{Department of Statistics and Data Sciences, University of Texas at Austin, USA}
\icmlaffiliation{queen}{School of Mathematics and Physics, The University of Queensland}
\icmlaffiliation{sf}{Salesforce Research}
\icmlaffiliation{hcmus}{AISIA Lab, University of Science, Vietnam National University Ho Chi Minh City}
\icmlaffiliation{astar}{Institute for Infocomm Research (I$^2$R), A$^*$STAR, Singapore}

\icmlcorrespondingauthor{Quang Pham}{hqpham.2017@phdcs.smu.edu.sg}

\icmlkeywords{Machine Learning, ICML}

\vskip 0.3in
]



\printAffiliationsAndNotice{\icmlEqualLast}  

\begin{abstract}
Sparse mixture of experts (SMoE) offers an appealing solution to scale up the model complexity beyond the mean of increasing the network's depth or width. However, effective training of SMoE has proven to be challenging due to the representation collapse issue, which causes parameter redundancy and limited representation potentials. In this work, we propose a competition mechanism to address this fundamental challenge of representation collapse. By routing inputs only to experts with the highest neural response, we show that, under mild assumptions, competition enjoys the same convergence rate as the optimal estimator. We further propose CompeteSMoE, an effective and efficient algorithm to train large language models by deploying a simple router that predicts the competition outcomes. Consequently, CompeteSMoE enjoys strong performance gains from the competition routing policy while having low computation overheads. Our extensive empirical evaluations on two transformer architectures and a wide range of tasks demonstrate the efficacy, robustness, and scalability of CompeteSMoE compared to state-of-the-art SMoE strategies. 
\end{abstract}
\vspace{-0.3in}

\section{Introduction} \label{sec:intro}
Large language models (LLMs) have emerged as a promising framework for artificial general intelligence. In recent years, LLMs have shown a remarkable success in solving many cognitive tasks, ranging from language, vision understanding~\citep{bao_vlmo_2022, gulati_conformer_2020,dosovitskiy_image_2021, ruiz_scaling_2021, bao_beit_2022,li2022blip,li2023blip}, to code generation~\citep{wang2021codet5}, reinforcement learning~\citep{chow_mixture_expert_2023} and life sciences~\citep{rives_biological_2021}. Since the release of the original Transformer model~\citep{vaswani_attention_2017}, extensive efforts have been devoted to scaling up the model complexities to take advantage of massive datasets and advanced computing resources~\citep{radford2019language,brown2020language,du_glam_2022}. To go beyond simply increasing the network depth and width, Sparse Mixture-of-experts (SMoE)~\citep{fedus_switch_2022} has risen as an appealing solution for scaling LLMs. By modularizing the network and activating only subsets of experts per input, SMoE offers constant computational costs while scaling up the model complexity, which often results in improved performances.

Despite the initial success, effective SMoE training has been well-known to be notoriously challenging because of the representation collapse issue~\citep{lee_thorp_sparse_2022,riquelme2021scaling,chi_representation_2022} where all experts converge to learn similar representations or all tokens are only routed to a few experts. As a result, SMoE often suffers from limited representation capabilities and wasteful parameter usage. Thus, over the years, advances in SMoE algorithmic research are driven by the efforts to alleviate the representation collapse problem. Nevertheless, state-of-the-art strategies mostly rely on intuitive conceptualizations, which can only offer greedy solutions such as using a cosine router~\citep{chi_representation_2022} or a two-phase training procedure~\citep{dai_stablemoe_2022}. Furthermore, due to the large scale nature of LLM experiments, empirical evaluations are often restricted to the vanilla Transformer, which does not take into account the advanced architectures such as GLaM~\citep{du_glam_2022} or Brainformer~\citep{zhou2023brainformers}.

This work focuses on improving the training effectiveness of SMoE by addressing its core challenge of representation collapse. To this end, we first seek to develop a reliable strategy to route inputs to experts. 
Motivated by the Winner-take-all (WTA) principle~\cite{grossberg1982contour} originated from biology~\citep{riesenhuber1999hierarchical,andersen1969participation,eccles2013cerebellum}, we propose the \emph{competition} mechanism for SMoE training, which works by routing token only to experts with the highest neural responses. By doing so, experts are encouraged to \emph{compete} with one another to be associated to the current input and result in a natural and parameter-free routing algorithm.
We further investigate the competition's theoretical learning property by showing that it has \emph{the same} convergence rate as the \emph{optimal estimator in hindsight}. 
Based on this solid foundation, we develop \emph{CompeteSMoE} as a novel SMoE training framework for LLMs. CompeteSMoE innovatively employs a router trained to both minimize the task loss and predict the competition outcomes. As a result, CompeteSMoE produces high quality routing policy that are relevant to the task of interest with low computational overhead. We empirically compare CompeteSMoE with a suite of state-of-the-art SMoE learning strategies on two large-language-model architectures to demonstrate its efficacy, scalability, and robustness.

In summary, our work contributes the following innovations. First, we propose a novel \emph{competition} mechanism for training SMoE, which enjoys the same convergence rate as the optimal estimator in hindsight. Second, we develop \emph{CompeteSMoE}, a scalable and effective training strategy for SMoE training via competition. CompeteSMoE employs a router trained to predict the competition outcome in a scheduled manner. Thus, the router can learn high quality routing policy that are relevant to the current task. Lastly, we conduct extensive experiments to demonstrate strong learning capabilities of CompeteSMoE and show its promising scalability to large scale architectures.

\section{Background} \label{sec:background}
\begin{table*}[!h]
\centering
\setlength\tabcolsep{3.5pt}
\begin{tabular}{llcccc}
\toprule
Method & Venue & Router  & Router training & Routing objective & Guarantee \\ \midrule
Switch Transformer & JMLR 2022 & Linear & Yes & Greedy & No \\
XMoE & NeurIPS 2022 & Cosine & Yes & Greedy & No \\
StableMoE & ACL 2022 & &  & &  \\
\quad - Stage 1 & - & Linear & Yes &Greedy & No \\
\quad - Stage 2 & - & Linear & No & Fixed Routing & No \\
SMoE-Dropout & ICLR 2023 & Linear & No & Fixed Routing & No \\ \midrule
CompeteMoE & - & Linear & Scheduled & Competition & Yes \\ \bottomrule
\end{tabular}
\caption{Characteristics of state-of-the-art SMoE strategies for training LLMs.} \label{tab:char}
\vspace{-0.1in}
\end{table*}

Table~\ref{table_notations} in the Appendix summarizes all notations used in this work.
We first describe the SMoE training process, which consists of a router $\gR(\cdot,W_r)$ parameterized by $W_r$ and $N$ experts $\{g(\cdot,W_{e_i}) \}_{i=1}^N$ parameterized by $W_{e_i}, i \in [N]$, respectively.
We follow~\citet{fedus_switch_2022} and only implement SMoE in the fully connected layers.
Given an input token $\vx$ and its representation $\vz \in \mathbb{R}^d$, SMoE first uses its router to calculate the affinity scores between $\vz$ and each experts as $\vs = \gR(\vh)$. Then, a sparse gating function, $\mathrm{TopK}$, is applied to select only $K$ experts with the largest affinity scores. Here we define the $\mathrm{TopK}$ function as:
\begin{align} \label{eq:softmax}
    &\TopK(v_i,K)\nonumber\\
    &:=\begin{cases}
        ~v_i, \hspace{.32cm} \text{if } v_i \text{ is in the } K  \text{ largest elements of } v\\
        -\infty, \hspace{.15cm}\text{otherwise}.
    \end{cases}
\end{align}
Finally, the selected experts \emph{independently} calculate their outputs, which are then linearly combined according to their affinity scores as the final prediction as:
\begin{align} \label{eq:smoe}
    \hat{y} =& \sum_{i=1}^N \softmax(\mathrm{TopK}(s_i,i)) \cdot g(\vh; W_{e_i}),
\end{align}
where $\softmax(s_i):=\exp(s_i)/\sum_{j=1}^{K}\exp(s_j)$. In this work, we mainly focus on top-2 routing ($K=2$) since it has been shown to achieve the best trade-off between training efficiency and testing performance~\citep{lepikhin_gshard_2021,du_glam_2022,zhou2023brainformers}. 

The naive SMoE training in equation~(\ref{eq:smoe}) has been the standard practice in large scale implementations.
However, it is also susceptible to the representation collapse issue due to the joint optimization of the router and experts' parameter~\citep{lee_thorp_sparse_2022}. Therefore, extensive efforts have been devoted to improve the effectiveness of SMoE training by alleviating the representation collapse issue. The first line of work is based on the motivation that decoupling the router and expert training can help alleviate the collapse. Thus, StableMoE~\citep{dai_stablemoe_2022} proposes a two-stage training process where only the routers or experts are trained in each stage. Based on the similar idea, SMoE-Dropout~\citep{chen2023sparse} fixes a randomly initialized router and propose the ``self-slimmable" strategy that gradually increases the number of selected experts throughout training. Secondly, XMoE~\citep{chi_representation_2022} proposes to alleviate the collapse issue by implementing a deep router with a down-projection and normalization layers. 
{Then, HyperRouter~\citep{do_hyperrouter_2023} addresses the compromise between
fixed and trainable routers in SMoE training by leveraging a random but fixed Hypernetwork~\citep{ha_hypernetworks_2017} to dynamically generate the router parameters.}
However, such strategies either fix a random policy or train the router to minimize the task loss, which is often referred to as the greedy strategy~\citep{dai_stablemoe_2022}. Consequently, we argue that such greedy objectives could result in suboptimal routing policies and hinder the overall performances.

These limitations necessitate the need to develop a reliable strategy for selecting the most affinitive experts given an input and how to maintain this routing strategy throughout training. To this end, we propose CompeteSMoE with two key innovations: (i) using competition as the expert routing strategy; and (ii) a scheduled training process to distill the competition mechanism into a router, which results in both improve efficacy and low overheads. Table~\ref{tab:char} summarizes key characteristics of CompeteSMoE versus state-of-the-art SMoE training strategies discussed thus far. A detailed literature overview will be provided in Section~\ref{sec:related}.

\section{CompeteSMoE}
This Section details the competition mechanism and outlines the CompeteSMoE algorithm.

\subsection{Routing via Competition}
We propose the \emph{competition} mechanism as an effective routing policy. For simplicity, we consider a single fully connected layer implemented with the SMoE mechanism and leave the generalization to deep networks in Section~\ref{sec:competesmoe}. The competition mechanism's key innovation is the use of expert's activation norm as its affinity score, i.e. $s_i = ||g(\vh, W_{e_i})||_2$. Consequently, the competition-based SMoE training procedure can be summarize as: 
\begin{enumerate} 
    \item Calculate each expert output: $g(\vh, W_{e_i}), \; \forall i \in [N]$ 
    \item Calculate the expert's affinity score: \newline {$s_i = ||g(\vz, W_{e_i})||_2, \; \forall i \in [N]$}
    \item Calculate the final output: \newline $\hat{y} = \sum_{i=1}^K \mathrm{softmax}(\mathrm{TopK} (s_i,K)) \cdot g(\vz, W_{e_i})$
\end{enumerate}
The key idea of competition is associating the expert's affinity score with its actual output, i.e. \emph{experts that have higher responses are more likely to be associated to the current input}. This strategy starkly contrasts the standard SMoE implementation discussed in Section~\ref{sec:background} where the affinity score is calculated as the dot product between the input $\vz$ and the expert's embedding, i.e. columns of $W_r$, and only the few selected experts actually perform their calculation.  
Although using expert embedding is more efficient, it results in suboptimal routing policies because the embedding is detached from the expert's forward calculation. Therefore, competition provides a trade-off between efficiency and an optimal routing policy. We will investigate its theoretical guarantees in Section~\ref{sec:understanding}.   

\subsection{Scheduled Training the Router}
\textbf{Router loss.}\; One drawback of the competition mechanism is that it requires activating all experts to calculate the affinity scores, which is expensive in large models. Therefore, to facilitate the large scale integration of competition, we propose to employ a router $\gR(\cdot,W_r)$ described in Section~\ref{sec:background}. Notably, instead of greedy training, we propose to train the router to \emph{predict the competition outcomes}, which can be characterized by the mean-squared error (MSE) between the competition and router policies. Particularly, let $\vs_{\gR}$ and $\vs_C$ be the affinity scores from the router and the competition, respectively, we propose to train the router by minimizing the router loss, denoted by $\mathcal{L}_{\gR}(\vs_{\gR}, \vs_C)$, defined as follows:
\begin{align} \label{eq:router_loss}
    & \gL_{\gR_l}(\vs_{\gR}, \vs_{C}) := \\ \notag
    &\mathrm{MSE}({\mathrm{softmax}(\mathrm{TopK}(\vs_{\gR},K))}, {\mathrm{softmax}(\mathrm{TopK}(\vs_C,K))}).
\end{align}

\begin{figure*}[t]
    \centering
    \begin{subfigure}{0.6\textwidth}
         \centering
         \includegraphics[width=\textwidth]{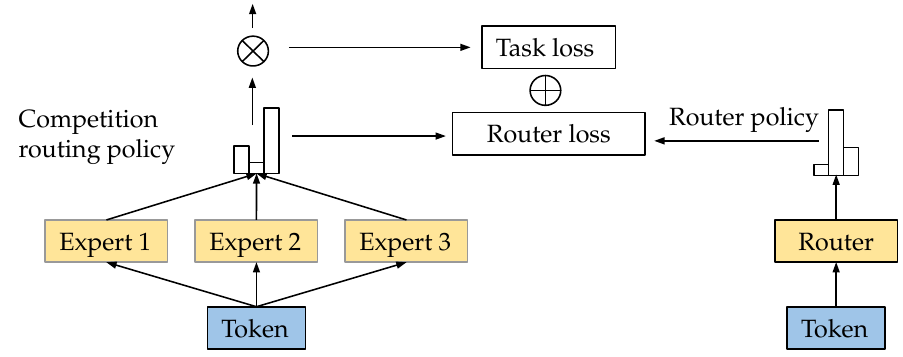}
         \caption{With probability $\lambda(t)$, train the router to minimize the competition outcomes and task loss}
         \label{fig:head}
     \end{subfigure}
     \hfill
     \begin{subfigure}{.38\textwidth}
         \centering
         \includegraphics[width=.8\textwidth]{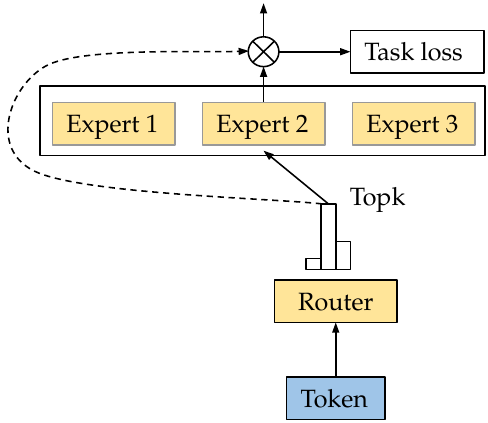}
         \caption{With probability $1-\lambda(t)$, train the routers and experts to minimize the task loss}
         \label{fig:tail}
     \end{subfigure}
     \hfill
     \caption{An illustrative of the CompeteSMoE algorithm on three experts.} \label{fig:algo}
     \vspace{-0.1in}
\end{figure*}

Consequently, we expect to use a well-trained router as a proxy to the competition routing policy without the expensive costs of activating all experts at every iteration.

\textbf{Scheduled training.}\; Due to the expensive computation, competition training of the router should only be performed sparingly. \citet{dai_stablemoe_2022} explored a two-phase training process where only the router is trained in the first phase and only the experts are trained in the second phase. However, since the experts are randomly initialized at the beginning, this strategy only fits the routing policy of random experts and does not model the experts' evolution when they observe training samples. Therefore, we propose a scheduled training strategy to ensure that the router can take into account the experts evolution.

To this end, at iteration $t$, we perform a coin flip with probability for head $\lambda(t)$ to decide whether to perform competition (head) or normal router training (tail). For competition training, the router is trained to minimize both the task loss (negative log-likelihood) and the router loss in equation~(\ref{eq:router_loss}).
By changing $\lambda(t)$, one can control how often the router can learn from the competition. In one extreme, by setting $\lambda(t) = 0$, we achieve the standard SMoE training. In this work, we simply implement $\lambda(t)$ as a small constant (e.g. $\lambda(t) = 0.05$). 
We explore and compare more different choices of $\lambda(t)$ in the numerical experiment and leave the more sophisticated designs of $\lambda(t)$ for the future work.

\subsection{The CompeteSMoE Algorithm} \label{sec:competesmoe}
We now introduce the CompeteSMoE algorithm that effectively trains large scale models via competition. CompeteSMoE focuses to facilitate SMoE training of LLMs, which comprise a stack of multihead self-attention (MHSA), fully-connected (FC), and  SMoE layers~\citep{fedus_switch_2022}. 

Given a network with $L$ layers, CompeteSMoE replaces each SMoE layer \textbf{independently} with its competition mechanism, which comprises a set of experts, a router, and a schedule. At iteration $t$, CompeteSMoE performs $L$ coin flips to determine the scheduled training at each layer. Then, CompeteSMoE updates the experts parameters to minimize the task loss while updating the routers according to their schedule. 
Figure~\ref{fig:algo} illustrates the CompeteSMoE algorithm for one layer of three experts.
In summary, CompeteSMoE training dynamic is outlined as: 
\begin{align}
    W_e^l \gets& W_e^l - \epsilon_t \frac{\partial}{\partial W_e^l} \gL_{\mathrm{NLL}} (\hat{y}, y), \quad l \in [L] \\
    W_r^l \gets& W_r^l - \epsilon_t \frac{\partial}{\partial W_r^l} \big[\gL_{\gR_l}(\vs_{\gR_l}, \vs_{C_l}) \nonumber\\
    & \hspace{2.2cm}+ \alpha \gL_{\mathrm{NLL}} (\hat{y}, y)\big], \quad l \sim \Lambda(t)
\end{align}
where $\gL_{\mathrm{NLL}}$ is the negative log-likelihood between the predicted output $\hat{y}$ and the ground-truth $y$, $\gL_{\gR}$ is the router loss defined in equation~(\ref{eq:router_loss}), $\epsilon_t$ is the current step size, $\alpha$ is the balance factor, and $\Lambda(t)$ is the set of all layers performing competition according to the schedule $\lambda(t)$.

We highlight two key ideas when implementing CompeteSMoE. First, the competition mechanism should be activated independently per layer rather than globally across the network. Although competition is guaranteed to find an optimal routing policy for a single SMoE layer (as we will show in Section~\ref{sec:understanding}), performing competition simultaneously in a deep network is a greedy strategy and does not guarantee an optimal routing policy across layers. We argue that performing competition independently per layer can improve the representation robustness and at a cheaper computation cost.
Secondly, the router optimization dynamic involves the interleaving of the router with the negative log-likelihood losses. While the router loss drives the routing policy toward optimal, the negative log-likelihood loss ensures that the learned policy is helpful in performing the current task.

\section{Statistical Guarantee of the Competition Mechanism}
\label{sec:understanding}


In this section, we investigate the convergence rate of the SMoE trained with competition and show that it enjoys the same convergence rate as the best estimator in hindsight.

\textbf{Problem setting.}
Let $(X_1,Y_1)\ldots,(X_n,Y_n) \in \gX^d \times \gY$ be i.i.d samples drawn from bounded subsets $\gX^d \subset \mathbb{R}^d$ and $\gY \subset \mathbb{R}$. We consider the following density function to characterize SMoE with the competition mechanism: 
\begin{align}
    p_{G_*}(Y|X)
    &:=\sum_{i=1}^{k_*}\Big[\softmax(\TopK(|g(x,W^*_{e_i})|,K))\nonumber\\
    &\hspace{2.5cm} \times f(Y|g(x,W^*_{e_i}),\sigma^*_i)\Big].
\end{align}
Here, $k*$ is a true number of experts and for any vector $v=(v_i)_{i=1}^{k_*}$, the softmax and TopK functions are defined in equations~(\ref{eq:softmax}) and~(\ref{eq:smoe}) respectively.
Additionally, $g(X,W_e)$ stands for an expert function, while $f(\cdot|\mu,\sigma)$ denotes an univariate Gaussian density with mean $\mu$ and variance $\sigma$. 
In addition, we also define ${G_*:=\sum_{i=1}^{k_*}\delta_{(W^*_{e_i},\sigma^*_i)}}$ as a true mixing measure where $(W^*_{e_i},\sigma^*_i)$ are ground-truth expert parameters and $\delta$ is the Dirac measure. In this work, we assume that $(W^*_{e_1},\sigma^*_1),\ldots,(W^*_{e_{k_*}},\sigma^*_{k_*})\in\Theta$ are pairwise different with $\Theta$ being a compact subset of $\mathbb{R}^q\times\mathbb{R}_+$ for some $q\in\mathbb{N}$.
Next, we assume that the expert function $g(X,W_e)$ is non-zero and differentiable with respect to $W_e$ for almost surely $X$. Furthermore, if there exist $\alpha_{u}\in\mathbb{R}$ for $1\leq u\leq q$ such that $\sum_{u=1}^{q}\alpha_u\frac{\partial g}{\partial W_e^{(u)}}(X,W_e)=0$ for almost surely $X$, then we must have $\alpha_u=0$ for all $1\leq u\leq q$.

\textbf{Maximum Likelihood Estimation.} In this paper, we propose using the maximum likelihood estimation (MLE) method to estimate unknown parameters in the above model as follows:
\begin{align}
    \label{eq:MLE}
    \widehat{G}_n\in\argmax_{G\in\mathcal{E}_{k_*}(\Theta)}\frac{1}{n}\sum_{i=1}^{n}\log(p_{G}(Y_i|X_i)),
\end{align}
where $\mathcal{E}_{k_*}(\Theta):=\{G=\sum_{i=1}^{k_*}\delta_{(W_{e_i},\sigma_i)}:(W_{e_i},\sigma_i)\in\Theta\}$ stands for the set of all mixing measures with exactly $k_*$ components. 

\begin{theorem}[Density Estimation]
    \label{theorem:density_estimation}
     With the MLE defined in~\eqref{eq:MLE}, the convergence rate of density estimation $p_{\widehat{G}_n}(Y|X)$ to the true density $p_{G_*}(Y|X)$ is given by:
    \begin{align*}
        \mathbb{P}\Big(\mathbb{E}_X[h(p_{\widehat{G}_n}(\cdot|X),p_{G_*}(\cdot|X))] &>C\sqrt{\log(n)/n}\Big)\nonumber\\
        & \lesssim\exp(-c\log(n)),
    \end{align*}
    for some universal positive constants $C$ and $c$ depending only on $\Theta$. Here, $h$ is the Hellinger distance defined as $h^2(f_1,f_2):=\frac{1}{2}\int(\sqrt{f_1}-\sqrt{f_2})^2\dint \nu$ for any two probability density functions $f_1,f_2$ dominated by the Lebesgue measure $\nu$.
\end{theorem}
Proof of Theorem~\ref{theorem:density_estimation} is in Appendix~\ref{appendix:density_estimation}. It follows from this theorem that $p_{\widehat{G}_n}$ converges to its true counterpart $p_{G_*}$ under the Hellinger distance at a parametric rate of order $\mathcal{O}(n^{-1/2})$ (up to some logarithmic term). Subsequently, we leverage this result to establish the estimation rates of ground-truth parameters $(W^*_{e_j},\sigma^*_j)$ for any $j\in[k_*]$.

Given some mixing measure $G:=\sum_{i=1}^{k_*}\delta_{(W_{e_i},\sigma_i)}$, we distribute its components to the following Voronoi cells which are generated by the support of the true mixing measure $G_*:=\sum_{j=1}^{k_*}\delta_{(W^*_{e_j},\sigma^*_j)}$:
\begin{align}
    \mathcal{A}_{j}&\equiv\mathcal{A}_{j}(G)\nonumber\\
    & :=\{i\in[k_*]:\|\theta_i-\theta^*_j\|\leq\|\theta_i-\theta^*_{\ell}\|, \ \forall \ell\neq j\}, \label{eq:Voronoi_cells}
\end{align}
where we denote $\theta_i:=(W_{e_i},\sigma_i)$ and $\theta^*_j:=(W^*_{e_j},\sigma^*_j)$ for any $i,j\in[k_*]$. 
Based on these Voronoi cells, we propose the following Voronoi loss functions for the sake of characterizing the convergence rates of parameter estimation:
\begin{align}
    \label{eq:Voronoi_loss}
    &\mathcal{D}(G,G_*)\nonumber\\
    &:=\max_{\{\tau_1,\tau_2,\ldots,\tau_K\}}\sum_{j=1}^{K}\sum_{i\in\mathcal{A}_{\tau_j}}\Big[\|W_{e_{i}}-W^*_{e_{\tau_j}}\|+|\sigma_{i}-\sigma^*_{\tau_j}|\Big].
\end{align}
Here, the maximum is subject to all $K$-element subsets $\{\tau_1,\tau_2,\ldots,\tau_K\}$ of $\{1,2,\ldots,k_*\}$.
\begin{theorem}[Parameter Estimation]
    \label{theorem:parameter_estimation}
    When the true number of experts $k_*$ is known, we show that the following Hellinger lower bound holds for any mixing measure $G\in\mathcal{E}_{k_*}(\Theta)$:
    \begin{align}
        \label{eq:Hellinger_lower_bound}
        \mathbb{E}_X[h(p_{G}(\cdot|X),p_{G_*}(\cdot|X))]\gtrsim \mathcal{D}(G,G_*).
    \end{align}
    This lower bound together with the density estimation rate in Theorem~\ref{theorem:density_estimation} leads to the following convergence rate of the MLE $\widehat{G}_n$ to the true mixing measure $G_*$:
    \begin{align}
        \label{eq:G_bound}
        \mathbb{P}\Big(\mathcal{D}(\widehat{G}_n,G_*)>C_1\sqrt{\log(n)/n}\Big)\lesssim\exp(-c_1\log(n)),
    \end{align}
    where $C_1$ and $c_1$ are some universal constants.
\end{theorem}
Proof of Theorem~\ref{theorem:parameter_estimation} is in Appendix~\ref{appendix:parameter_estimation}. Equation~(\ref{eq:G_bound}) indicates that the MLE $\widehat{G}_n$ also converges to $G_*$ at the parametric rate of order $\mathcal{O}(n^{-1/2})$ under the Voronoi metric $\mathcal{D}$. As a consequence, the rates for estimating both $W^*_{e_i}$ and $\sigma^*_i$ are optimal at $\mathcal{O}(n^{-1/2})$ for any $i\in[k_*]$.

\section{Related Work} \label{sec:related}
\subsection{Sparse Mixture of Experts}
Mixture of Experts (MoE) is a fundamental model in machine learning~\citep{jacobs_adaptive_1991,jordan_hierarchical_1994} and an instance of the conditional computation framework where different experts are responsible for different regions of the input space~\citep{yuksel_twenty_2012,bengio_deep_2013,masoudnia_mixture_2014,nguyen_practical_2018,nguyen_model_2021}. In literature, extensive efforts have been devoted to establishing a theoretical foundation for MoE, including the universal approximation properties~\citep{norets_approximation_2010,nguyen_universal_2016,nguyen_approximation_2019,nguyen_approximation_2020,nguyen_approximations_2021,nguyen_approximation_2023}, model selection criterion~\citep{khalili_new_2010,montuelle_mixture_2014,nguyen_l_1_oracle_2021,nguyen_non_asymptotic_2022,nguyen_non_asymptotic_2023}, convergence rate for density estimations~\citep{mendes_convergence_2012,norets_adaptive_2021,norets_adaptive_2022} and the problem of parameter estimation~\citep{ho_convergence_2022,nguyen_demystifying_2023,nguyen2024gaussian, Nguyen2024temperature}.

Investigating MoE to handle complex data structures such as vision or speech is an equally attractive research direction~\citep{eigen_learning_2013}. However, since activating all experts for every input is expensive for large models, a sparse version~\citep{shazeer_outrageously_2017} (SMoE) is often more attractive since it offers low computational costs while enjoying improved representation capabilities. To this end, SMoE has been widely explored to improved the training efficiency of LLMs, with various routing strategies proposed, from (i) letting tokens select the top-$k$ experts~\citep{lepikhin_gshard_2021,fedus_switch_2022,zuo_taming_2022,chi_representation_2022,dai_stablemoe_2022,chen2023sparse}, (ii) letting experts select the top-$k$ tokens~\citep{zhou_mixture_experts_2022}, to (iii) globally decide expert assignment \citep{lewis_base_2021,clark_unified_2022}. 
Nevertheless, despite the empirical success, advanced in theoretical studies~\citep{nguyen_demystifying_2023,nguyen2024statistical, Nguyen2024temperature} have not been translated into practical and effective algorithms for large scale modes.
In contrast, our work goes beyond both the pure theoretical or analytical studies by developing a theoretically-grounded algorithm for effective training of large scale LLM models

\subsection{Competitive Learning}
Competitive learning refers to a framework where computational units compete with one another for the right to response to an input~\citep{mcclelland1987parallel}. Its development is closely related to the biological brain where only certain cells respond strongly to a particular pattern and send suppressive signals to the remaining cells~\citep{andersen1969participation,stefanis1969interneuronal,eccles2013cerebellum}. 
Early investigations of competitive learning showed encouraging results in various domains such as action selection~\citep{feldman1982connectionist}, self-organizing maps~\citep{von1973self,kohonen1982self}, feature discovery~\citep{rumelhart1985feature}, and spiking networks~\citep{oster2005spiking}. Recently, the competition mechanism also motivates the development of various advanced machine learning methods such as maxout networks~\citep{goodfellow2013maxout}, compete to compute~\citep{srivastava2013compete}, and independent mechanisms~\citep{goyal_recurrent_2021,alias2021neural}. The work most related to us is TIM~\citep{lamb2021transformers}, which uses competition as an inductive bias to the vanilla Transformer and implements it via the MHSA mechanisms. However, this study shares the same weaknesses as the competitive learning framework in that it always activates all experts, which makes scaling to million parameters prohibitive. In contrast, our work not only provides a theoretical guarantee for the competition mechanism but also a training framework to distill this mechanism to a simple router and enjoy the effective routing policy with low overheads.

\section{Experiment} \label{sec:exp}
We conduct experiments to investigate the following hypotheses: (i) CompeteSMoE offers an effective SMoE training algorithm for LLMs; and (ii) CompeteSMoE scales well with the model complexity, computational resources, while being robust to the hyper-parameter configurations. 

\subsection{Experimental Settings}
\textbf{Training tasks}\; We explore two common tasks in training and evaluation of LLMs. First, character-level language modeling on the enwik8 or text8 datasets~\citep{mahoney_large_2011}, which are common datasets to evaluate the model's pre-training capabilities. We also consider the word-level language modeling task on WikiText-103~\citep{merity_pointer_2017}, a much larger and more challenging dataset, to test the models scaling capabilities. For all datasets, we follow the default splits of training-validation-testing. Second, we consider finetuning the models on downstream applications to investigate the models capabilities of adapting to different domains. To this end, we consider pre-trained medium models on enwik8 and finetuning it on a downstream task. We choose the SST-2~\citep{socher_recursive_2013}, SST-5~\citep{socher_recursive_2013}, IMDB~\citep{maas_learning_2011}, and BANKING77~\citep{casanueva-etal-2020-efficient} datasets, which are common NLP tasks to evaluate pre-trained models. Following~\citet{chen2023sparse}, we freeze the router and only optimize the experts' parameter in this experiment.

\begin{table*}[!htbp]
\centering
\begin{tabular}{@{}llcccccc@{}}
\toprule
\multicolumn{2}{c}{Configuration}                & \multicolumn{2}{c}{Enwik8 (BPC)} & \multicolumn{2}{c}{Text8 (BPC)} & \multicolumn{2}{c}{WikiText-103 (Perplexity)} \\ \midrule
Architecture                 & Algorithm   & Small            & Medium  & Small       & Medium      & Small           & Medium         \\ \midrule
\multirow{5}{*}{Switch Transformer} & CompeteSMoE & \textbf{1.177}   & \textbf{1.115}   & \textbf{1.289}       & \textbf{1.220}       & \textbf{83.750}          &       \textbf{31.299}         \\
                             & SMoE        & 1.191            & 1.130   & 1.291       & 1.231       & 85.061          & 34.145         \\
                             & SMoE-Fixed  & 1.202            & 1.138   & 1.308       & 1.237       & 86.656          & 35.392         \\
                             & XMoE        & 1.198            & 1.126   & 1.291       & 1.244       & 84.686          & 34.322         \\
                             & StableMoE   & 1.194            & 1.124   & 1.300       & 1.238       & 85.267          & 34.179         \\ \midrule
\multirow{5}{*}{GLaM}        & CompeteSMoE & \textbf{1.175}            & \textbf{1.081}   & \textbf{1.281}       & \textbf{1.211}       &   \textbf{54.380}              &     \textbf{34.233}           \\
                             & SMoE        & 1.186            & 1.139   & 1.286       & 1.235       &    54.856             &       34.672         \\
                             & SMoE-Fixed  & 1.191            & 1.123   & 1.308       & 1.238       &  57.284               &         35.640       \\
                             & XMoE        & 1.187            & 1.111   & 1.289       & 1.219       &       55.618          &       34.613         \\
                             & StableMoE   & 1.189            & 1.109   & 1.291       & 1.217       &    54.730             &     35.320   
                             \\ \bottomrule
\end{tabular}
\caption{BPC on the enwik-8 and text8 test sets; and perplexity on the Wikitext-103 test set. Lower is better, best results are in bold.} \label{table:pre-train}
\end{table*}

\textbf{Architecture.}
We consider two decoder-only architectures: (i) the standard Switch Transformer~\cite{fedus_switch_2022}; and (ii) and {GLaM}~\citep{du_glam_2022}, a more advanced SMoE architecture. Due to resource constraints, training massive LLM models is prohibitive without industrial resources. Thus, we consider three model configurations: (i) tiny: with three SMoE blocks and \textbf{7M} parameters; (ii) small: with six SMoE layers and \textbf{15M} parameters; and (iii) medium: with eight SMoE layers and \textbf{36M} parameters. We emphasize that we are not trying to achieve state-of-the-art results due to the limited resource constraints. Instead, we evaluate the small and medium models on various datasets to demonstrate the scalability and efficacy of our algorithm. Lastly, we conduct extensive investigations using the tiny model to understand the algorithm behaviours and its robustness to different design choices. Lastly, unless otherwise stated, we implement $\TopK$ with $K=2$ in the experiments.


\textbf{Baselines.}
We compare our CompeteSMoE with state-of-the-arts SMoE training strategies for LLMs. \textbf{SwitchTransformer}~\citep{fedus_switch_2022} employs a simple router trained end-to-end with the experts. \textbf{StableMoE}~\cite{dai_stablemoe_2022} proposes a two-phase training process where the first phase greedily trains only the router, then the router is fixed to train the experts in second phase. \textbf{XMoE}~\cite{chi_representation_2022} implements a deep router that comprises a down-projection and normalization layer and a gating network with learnable temperatures. Lastly, motivated by {SMoE-Dropout}~\cite{chen2023sparse}, we implement the \textbf{SMoE-Fixed} strategy that employs a randomly initialized router and freeze it throughout the training process. We do not consider gradually activate all experts since it is infeasible for large scale models.

\textbf{Training procedure.}\; 
For the language modeling experiments, we optimize the small models for 60,000 steps and the medium models for 80,000 steps. We use an Adam~\citep{kingma2014adam} optimizer with a learning rate schedule of $1/\sqrt{t}$. The lowest validation loss checkpoint is used to report the final performance on the test set. 
For XMoE and StableMoE, we follow the default hyper-parameter configurations. For CompeteSMoE, there are two hyper-parameters: the schedule $\lambda(t)$ and the balance factor $\alpha$. We first cross-validate the schedule $\lambda(t)$ with a random value of $\alpha$. Then, we proceed to cross-validate $\alpha$ with respect to the optimal $\lambda(t)$ found. This strategy's complexity only grows linearly with the number of configurations, which is more suitable for large models with million parameters compared to other alternatives.
For each finetuning dataset, we consider the pretrained checkpoint of the medium model on enwik8, remove the last layer and employ two randomly initialized fully connected layers as the classifier. We finetune all methods for 5,000 steps with the same learning rate. Lastly, all experiments do not consider the balancing loss~\citep{fedus_switch_2022} so that we can focus on validating the routing policy quality. We will leave the extension of competition and the balancing loss to the future work. Lastly, we run the the tiny configuration five times and only one time for other configurations because of the large scale nature of these experiments. Due to space constraints. we provide all experiment details, including hardware setup, standard deviations, and additional visualization in the Appendix.

\subsection{Language Modeling Evaluation}\label{sec:pretrain}
\begin{figure}[t]
    \includegraphics[width = 0.46\textwidth]{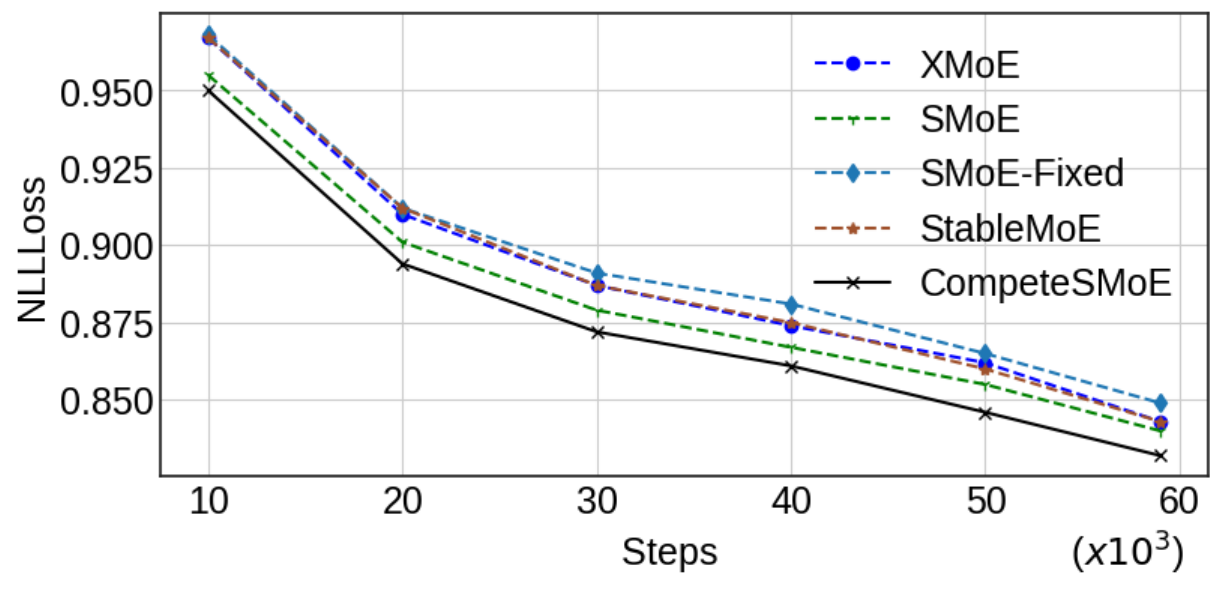}
     \caption{Validation loss of the small transformer model on enwik8 throughout training.} \label{fig:evolution_valiation}
     \vspace{-0.2in}
\end{figure}
Table~\ref{table:pre-train} reports the evaluation metrics of CompeteSMoE versus state-of-the-art strategies. We also report the evolution of the performance on the validation set of the small model in Figure~\ref{fig:evolution_valiation}.
We first observe that across all methods, the GLaM architecture offers better performances than the standard Switch Transformer on all datasets. Furthermore, state-of-the-art strategies like XMoE or StableMoE tend to outperform the vanilla SMoE when we increase the model complexity (from small to medium) or introduce more data (from enwik8 to WikiText-103). However, the improvements of these strategies are quite inconsistent or marginal. In contrast, our CompeteSMoE can consistently outperform other competitors on all benchmarks (note that the BPC metric is log-scaled), architectures, and offer faster convergent rate (Figure~\ref{fig:evolution_valiation}). This result demonstrates CompeteSMoE's capabilities in learning a good routing policy to facilitate the language modeling pre-training task.

Table~\ref{table:pre-train} also shows an interesting result that SMoE-Fixed performs the worse among all methods considered, showing that the random policy is insufficient and needs to be updated for effective training. This result sheds light on the success of SMoE-Dropout~\citep{chen2023sparse} by showing its performance gains mostly come from the self-slimmable strategy of linearly increasing the number experts activated ($K$) throughout training. However, self-slimmable transforms the sparse network into dense, which goes against the motivation of SMoE for large scale models.

\subsection{Finetuning Evaluation}
\begin{table}[!ht]
\resizebox{\linewidth}{!}{%
\begin{tabular}{@{}lcccc@{}}
\toprule
Method            & {SST-2} & {SST-5} & {IMDB} & {BANKING77} \\ \midrule
CompeteSMoE & \textbf{79.8}             & \textbf{38.3}             & \textbf{84.8}            & \textbf{72.9}                 \\
SMoE        & 77.1                      & 35.1                      & 84.4                     & 69.2                          \\
SMoE-Fixed  & 78.6                      & 34.4                      & 83.5                     & 66.7                          \\
XMoE        & 76.7                      & 35.3                      & 83.3                     & 67.4                          \\
StableMoE   & 77.7                      & 34.3                      & 83.9                     & 60.8                          \\ \bottomrule
\end{tabular}}
\caption{Accuracy of the model after finetuned on various datasets. Higher is better, best results are in bold.} \label{table:finetune}
\end{table}

Table~\ref{table:finetune} reports the accuracy of the models finetuned on the test sets of various datasets. Overall, we observe that CompeteSMoE demonstrates strong transfer learning capabilities by achieving the highest accuracy on all datasets. Notably, on the more challenging datasets of SST-5 and BANKING77, which have fewer training samples or more classes, we observe larger performance gains from CompeteSMoE versus the remaining baselines (over $3\%$ improvements compared to the second best method). This result shows that CompeteSmoE can learn routing policies that are not only good for pre-training but also exhibit strong transfer capabilities to various downstream tasks. 

\subsubsection{Scalability to Different $\TopK$ Values}
\begin{figure}[t]
    \includegraphics[width = 0.46\textwidth]{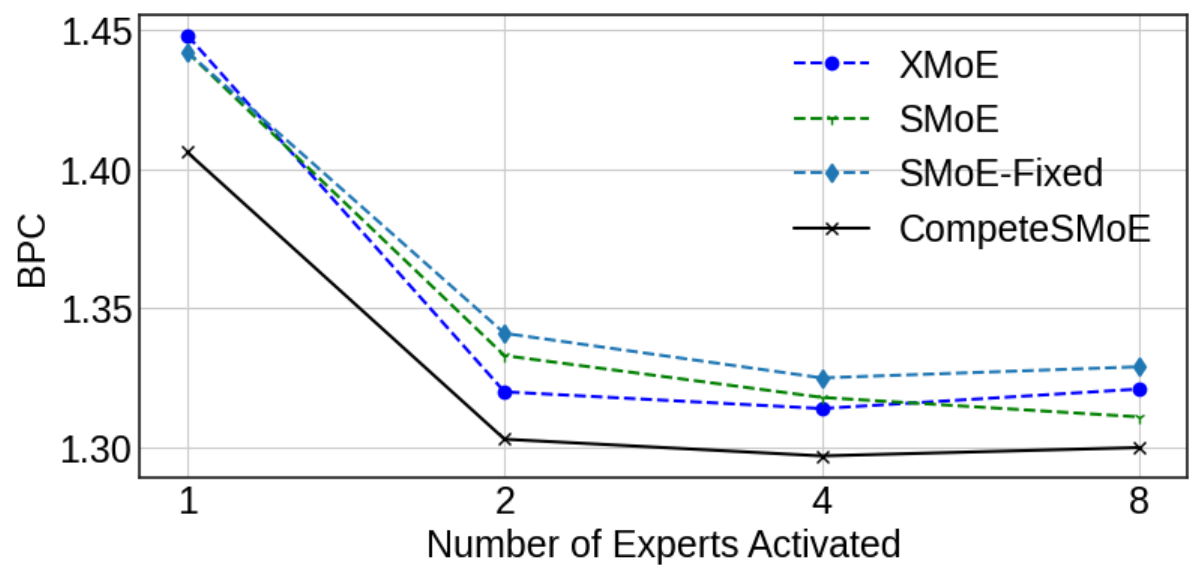}
     \caption{BPC on enwik8 wrt the number of experts activated.} \label{fig:scalability_K}
     \vspace{-0.1in}
\end{figure}
Figure \ref{fig:scalability_K} reports the BPC on enwik8 of various methods with respect to the number of experts. We only consider the Tiny transformer configuration for comparison. From Figure \ref{fig:scalability_K},  we first observe that SMoE-Fixed achieved the worst results in all cased, further showing the low quality routing policy from the randomly initialized routers. Overall, CompeteSMoE consistently outperforms SMoE and SMoE-Fixed across all values of $K$, showing its scalability to the number of experts activated.

It is important to note that when all experts are activated $K=16$, the predictions of each method are still different because of the different affinity scores from the router. Thus, this result shows that even when we fix the same experts, CompeteSMoE can learn to adjust the contribution of each experts accordingly to improve the learning outcomes.

\subsection{Ablation Study} \label{sec:ablation}

We investigate the robustness of CompeteSmoE to the different hyper-parameter settings. All experiments are conducted using the tiny transformer architecture. 
\subsubsection{Robustness to the schedule $\lambda(t)$}
\begin{table}[!ht]
\resizebox{\linewidth}{!}{%
\begin{tabular}{@{}lccccc@{}}
\toprule
Frequency $\lambda(t)$           & 0.01 & 0.03 & 0.05 & 0.07 & 0.09 \\ 
\midrule
SwitchTransformer  & 1.733 & 1.330 & \textbf{1.303} & 1.310 & 1.309 \\
GLaM & 1.587 & 1.366 & 1.298 & 1.300 & \textbf{1.294} \\
\bottomrule
\end{tabular}}
\caption{BPC on enwik8 with respect to $\lambda(t)$. Lower is better, best results are in bold.} \label{table:ablations_chedule}
\vspace*{-0.05in}
\end{table}
We first fix the the balance factor $\alpha$ and investigate the effect of the schedule $\lambda(t)$. Table \ref{table:ablations_chedule} reports the BPC on enwik8 with respect to $\lambda(t)$. When $\lambda(t)=0$, CompeteSmoE reduces to the vanilla SMoE. We first observe that when $\lambda(t)=0.01$, the competition learning is too sparse, and may inject noise to the learning process, resulting in the performance degrade. Once we increase the competition frequency, CompeteSMoE consistently outperforms SMoE, demonstrating the benefits of competition to the routing policy. Importantly, we observe that when $\lambda(t) \ge 0.05$, CompeteSMoE's performance becomes saturated, indicating that even reasonably small competition frequency is sufficient to achieve significant improvements. Furthermore, it is important to note that the balancing factor $\alpha$ is cross-validated with respect to $\lambda(t)=0.05$, which explains why it achieves the best performance in this experiment.

\subsubsection{Robustness to the balancing factor $\alpha$}
\begin{table}[!ht]
\resizebox{\linewidth}{!}{%
\begin{tabular}{@{}lccccc@{}}
\toprule
$\alpha$           & 0.1 & 1 & 5 & 25 & 250 \\ 
\midrule
SwitchTransformer  & 1.306 & 1.307 & \textbf{1.303} & 1.321 & 1.328 \\
GLaM & \textbf{1.280} & 1.285 & 1.298 & 1.304 & 1.319 \\
\bottomrule
\end{tabular}}
\caption{BPC on enwik8 with respect to the balancing factor $\alpha$. Lower is better, best results are in bold.} \label{table:ablations_balancing_factor}
\vspace*{-0.1in}
\end{table}
Lastly, we investigate the robustness of the balancing factor $\alpha$ given a fixed schedule $\lambda(t)$ and report the result in Table~\ref{table:ablations_balancing_factor}. Overall, we see that CompeteSmoE is robust to small values of $\alpha$, i.e., where $\alpha \le 5$. Such values provide a good trade-off between enforcing the router to follow the competition policy versus improving the learned policies for the current task, resulting in good overall performances. It is important to note that any $\alpha$ configurations reported offer better performances than SMoE ($\alpha=0$).

\textbf{Summary and hyper-parameter configuration guidelines.} Our experiments show that CompeteSMoE demonstrates not only strong pre-training and transfer learning capabilities but also robustness to the hyper-parameter configurations. Our ablation studies in Section~\ref{sec:ablation} suggest CompeteSMoE is robust to small values of the hyper-parameters, i.e., $0.03 \leq \lambda(t) \leq 0.09$ and $\alpha \leq 5$. This result serves as a general guide for efficiently cross-validating CompeteSMoE. We provide visualizations of the routers quality in Appendix~\ref{app:visualization}.

\vspace*{-0.1in}
\section{Conclusion}
This work proposes competition, a theoretically-grounded strategy for effective SMoE training. By routing tokens only to the winning experts, we show that competition enjoys the same convergence rate as the optimal routing policy in hindsight. From this foundation, we propose the CompeteSMoE algorithm that employs a router trained to predict the competition outcome. Consequently, CompeteSMoE enjoys the guarantee of competition while avoiding the expensive costs of activation all experts. We conduct extensive evaluations with two transformer architectures on both pre-training and finetuning tasks to demonstrate CompeteSMoE strong capabilities, scalability, and robustness compared to state-of-the-art training strategies. 
Lastly, we believe that extending the competition to train with the balancing loss and large scale evaluations of CompeteSMoE are two promising future research directions.

\section*{Impact Statements}

This paper presents work that aims to advance the field of Machine Learning. Our work is conducted in an academic setting using publicly available benchmarks and does not involve human subjects or proprietary assets.  Therefore, we believe that there are no significant societal implications of our work. However, since we have trained large language models from datasets collected from the web, there may be potential issues such as gender or racial biases, which will require additional efforts to mitigate their negative impact. In addition, despite encouraging results, training large numbers of LLMs is inevitably costly and requires substantial computing resources.

\bibliography{main}
\bibliographystyle{icml2024}

\newpage
\appendix
\onecolumn

\begin{center}
{\bf{\Large{Supplementary Material for ``CompeteSMoE - Effective Training of Sparse Mixture of Experts via Competition"}}}
\end{center}

This document is organized as follow. Appendix~\ref{app:notation} summarizes the notations used in our work. Appendix~\ref{app:proof} presents the detailed proof of our theoretical analysis in Section~\ref{sec:understanding}. Appendix~\ref{app:add_exp} presents all the implementation details, additional results, and visualizations.

\section{Summary of Main Notations} \label{app:notation}

We summarize the main notations used in the main paper in \cref{table_notations}, including those introduced later in the supplementary material.

\begin{table}[!th]
\centering
\begin{tabular}{cl}
\toprule
Symbol & Description \\ 
\midrule
$W_r$ & Router parameter \\
$W_{e}$ & Expert parameter\\
$N$ & Number of experts used \\ 
$\mathcal{R}$ & Router network (function)\\
$g$ & Expert network (function)\\
$\vx,~X$ & Token, input\\
$\vz$ & Token representation\\
$\vs$ & Affinity score\\
$\mathrm{TopK}$ & TopK function\\
$K$ & Top $K$ experts with the largest  affinity scores\\
$\widehat{y},~Y$ & Final prediction, output\\
$\vs_{\gR}$ & Router affinity score\\
$\vs_C$ & Competition affinity score\\
$t$ & Current $t$-th iteration\\
$\lambda(t)$ & Schedule (probability for head) at $t$-th iteration\\
$\gL_{\mathrm{NLL}}$ & Negative log-likelihood\\
$\gL_{\gR}$ & Router loss\\
$\epsilon_t$ & Step size\\
$\alpha$ & Balance factor\\
$n$ & Sample size\\
$[n]$ & Set of $\{1, 2, . . . , n\}$ for any positive integer $n$\\
$a_n = \mathcal{O}(b_n)$ or $a_{n} \lesssim b_{n}$ & If $a_n \leq C b_n$ for all $ n\in\mathbb{N}$, where $C > 0$ is some universal constant\\
$u^z$ & $u^z=u_{1}^{z_{1}}u_{2}^{z_{2}}\ldots u_{d}^{z_{d}}$, for any vector $u \in \mathbb{R}^{d}$ and $z:=(z_1,z_2,\ldots,z_d)\in\mathbb{N}^d$\\
$|u|$ & $|u|:=u_1+u_2+\ldots+u_d$,  for any vector $u \in \mathbb{R}^{d}$\\
$z!$ &$z!:=z_{1}!z_{2}!\ldots z_{d}!$, for any vector $u \in \mathbb{R}^{d}$ and $z:=(z_1,z_2,\ldots,z_d)\in\mathbb{N}^d$\\
$k^*$ & True number of experts\\
$f(\cdot|\mu,\sigma)$ &Univariate Gaussian density with mean $\mu$ and variance
$\sigma$\\
$G_*$ & True mixing measure\\
$\delta$ & Dirac measure\\
$\nu$ & Lebesgue measure\\
$\Theta$ & Parameter space\\
$q$ & Dimension of expert parameter space\\
$\widehat{G}_n$ & Maximum likelihood estimation of the true mixing measure $G_*$\\
$\mathcal{E}_{k_*}(\Theta)$ & Space of all mixing measures with exactly $k_*$ components\\
$\|\cdot\|,~\|\cdot\|_1$ & $2$-norm and $1$-norm value\\ 
$|A|$&
Cardinality of any set $A$\\
$h$ & Hellinger distance\\
$V$ & Total Variation distance\\
$\mathcal{D}$ & Voronoi loss function\\
$\mathcal{P}_k(\Theta),~\overline{\mathcal{P}}_k(\Theta),  ~\overline{\mathcal{P}}^{1/2}_{k_*}(\Theta)$ & Set of all conditional densities with respect to mixing measures in $\mathcal{E}_{k_*}(\Theta)$ and two variants\\
$\overline{\mathcal{P}}^{1/2}_{k_*}(\Theta,\delta)$ & Hellinger ball centered around the true conditional density $p_{G_*}(Y|X)$ and intersect with $\overline{\mathcal{P}}^{1/2}_{k_*}(\Theta)$\\
$\mathcal{J}_B(\delta,\overline{\mathcal{P}}^{1/2}_{k_*}(\Theta,\delta))$& Size of the Hellinger ball $\overline{\mathcal{P}}^{1/2}_{k_*}(\Theta,\delta)$\\
$N(\varepsilon,\mathcal{P}_{k_*}(\Theta),\|\cdot\|_1)$ & Covering number\\
$H_B(\varepsilon,\mathcal{P}_{k_*}(\Theta),h)$& Bracketing entropy\\ 
\bottomrule 
\end{tabular}
\caption{Summary of Main Notations.}
\label{table_notations}
\end{table}

\section{Proof for Results in Section~\ref{sec:understanding}} \label{app:proof}
\subsection{Proof of Theorem~\ref{theorem:density_estimation}}
\label{appendix:density_estimation}
In this proof, we first present some fundamental results on the density estimation problem for M-estimators in \cite{Vandegeer-2000}, and then provide the main proof in Appendix~\ref{appendix:main_proof}. To streamline our discussion, let us introduce some necessary concepts from the empirical process theory. In particular, let $\mathcal{P}_k(\Theta)$ be the set of all conditional densities with respect to mixing measures in $\mathcal{E}_{k_*}(\Theta)$, i.e.
\begin{align*}
    \mathcal{P}_{k_*}(\Theta):=\{p_{G}(Y|X):G\in\mathcal{E}_{k_*}(\Theta)\}.
\end{align*}
Additionally, we also consider two following variants of the set $\mathcal{P}_{k_*}(\Theta)$:
\begin{align*}
    \overline{\mathcal{P}}_k(\Theta)&:=\{p_{(G+G_*)/2}(Y|X):G\in\mathcal{E}_{k_*}(\Theta)\},\\
    \overline{\mathcal{P}}^{1/2}_{k_*}(\Theta)&:=\{p^{1/2}_{(G+G_*)/2}(Y|X):G\in\mathcal{E}_{k_*}(\Theta)\}.
\end{align*}
Next, we define for each $\delta>0$ a Hellinger ball centered around the true conditional density $p_{G_*}(Y|X)$ and intersect with the set $\overline{\mathcal{P}}^{1/2}_{k_*}(\Theta)$ as below
\begin{align*}
    \overline{\mathcal{P}}^{1/2}_{k_*}(\Theta,\delta):=\{p^{1/2}(Y|X)\in\overline{\mathcal{P}}^{1/2}_{k_*}(\Theta):h(p_{G},p_{G_*})\leq\delta\}.
\end{align*}
Moreover, the size of this Hellinger ball is quantified by the following term:
\begin{align}
    \label{eq:integral}
    \mathcal{J}_B(\delta,\overline{\mathcal{P}}^{1/2}_{k_*}(\Theta,\delta)):=\int_{\delta^2/2^{13}}^{\delta}H_B^{1/2}(t,\overline{\mathcal{P}}^{1/2}_{k_*}(\Theta,t),\|\cdot\|)\dint t\vee\delta,
\end{align}
where $H_B(t,\overline{\mathcal{P}}^{1/2}_{k_*}(\Theta,t),\|\cdot\|)$ stands for the bracketing entropy of $\overline{\mathcal{P}}^{1/2}_{k_*}(\Theta,t)$ under the $\ell_2$-norm, and $t\vee\delta:=\max\{t,\delta\}$. Now, we are ready to recall the results in \cite{Vandegeer-2000}. 
\begin{lemma}[Theorem 7.4,\cite{Vandegeer-2000}]
    \label{lemma:vandegeer}
    Take $\Psi(\delta)\geq\mathcal{J}_B(\delta,\overline{\mathcal{P}}^{1/2}_{k_*}(\Theta,\delta))$ such that $\Psi(\delta)/\delta^2$ is a non-increasing function of $\delta$. Then, for a universal constant $c$ and $\sqrt{n}\delta^2_n\geq c\Psi(\delta_n)$, we achieve that
    \begin{align*}
        \mathbb{P}\Big(h(p_{\widehat{G}_n},p_{G_*})>\delta\Big)\leq c\exp(-n\delta^2/c^2),
    \end{align*}
    for any $\delta\geq\delta_n$.
\end{lemma}
Proof of Lemma~\ref{lemma:vandegeer} is available in \cite{Vandegeer-2000}. Apart from this result, we also need to introduce the upper bounds of the covering number $N(\varepsilon,\mathcal{P}_{k_*}(\Theta),\|\cdot\|_1)$ and the bracketing entropy $H_B(\varepsilon,\mathcal{P}_{k_*}(\Theta),h)$ as follows:
\begin{lemma}
    \label{lemma:upper_bounds}
    Suppose that $\Theta$ is a bounded set, then we have for any $\varepsilon\in(0,1/2)$ that
    \begin{itemize}
        \item[(a)] $\log N(\varepsilon,\mathcal{P}_{k_*}(\Theta),\|\cdot\|_1)\lesssim\log(1/\varepsilon)$;
        \item[(b)] $H_B(\varepsilon,\mathcal{P}_{k_*}(\Theta),h)\lesssim\log(1/\varepsilon)$. 
    \end{itemize}
\end{lemma}
Proof of Lemma~\ref{lemma:upper_bounds} is deferred to Appendix~\ref{appendix:upper_bounds}.
\subsubsection{Main Proof}
\label{appendix:main_proof}
From \eqref{eq:integral} and part (b) of Lemma~\ref{lemma:upper_bounds}, we have that
\begin{align*}
    \mathcal{J}_B(\delta,\overline{\mathcal{P}}^{1/2}_{k_*}(\Theta,\delta))&=\int_{\delta^2/2^{13}}^{\delta}H_B^{1/2}(t,\overline{\mathcal{P}}^{1/2}_{k_*}(\Theta,t),\|\cdot\|)\dint t\vee\delta\\
    &\leq\int_{\delta^2/2^{13}}^{\delta}H_B^{1/2}(t,\overline{\mathcal{P}}^{1/2}_{k_*}(\Theta,t),h)\dint t\vee\delta\\
    &\lesssim\int_{\delta^2/2^{13}}^{\delta}\log(1/t)\dint t\vee\delta.
\end{align*}
Next, we choose $\Psi(\delta)=\delta\sqrt{\log(1/\delta)}$ such that $\Psi(\delta)\geq\mathcal{J}_B(\delta,\overline{\mathcal{P}}^{1/2}_{k_*}(\Theta,\delta))$. Then, with $\delta_n=\mathcal{O}(\sqrt{\log(n)/n})$, it follows from Lemma~\ref{lemma:vandegeer} that for some universal constants $C$ and $c$ which depend only on $\Theta$, we get
\begin{align*}
    \mathbb{P}\Big(h(p_{\widehat{G}_n},p_{G_*})>C\sqrt{\log(n)/n}\Big)\lesssim\exp(-c\log(n)).
\end{align*}
Hence, the proof is completed.

\subsubsection{Proof of Lemma~\ref{lemma:upper_bounds}}
\label{appendix:upper_bounds}
\textbf{Part (a).} Recall that $\Theta$ is a compact set, then there exists an $\varepsilon$-cover, which we denote as $\overline{\Theta}_{\varepsilon}$. Moreover, it can be verified that $|\overline{\Theta}_{\varepsilon}|\leq\mathcal{O}(\varepsilon^{-(q+1)k_*})$. Next, for each mixing measure $G=\sum_{i=1}^{k_*}\delta_{(W_{e_i},\sigma_i)}\in\mathcal{E}_{k_*}(\Theta)$, we consider another one $\overline{G}=\sum_{i=1}^{k_*}\delta_{(\overline{W}_{e_i},\overline{\sigma}_i)}$, where $(\overline{W}_{e_i},\overline{\sigma}_i)\in\overline{\Theta}_{\varepsilon}$ is the closet point to $(W_{e_i},\sigma_i)$ in this set for any $i\in[k_*]$. Then, the conditional density $p_{\overline{G}}(Y|X)$ belongs to the following set
\begin{align*}
    \mathcal{Q}:=\Big\{p_{G}(Y|X)\in\mathcal{P}_{k_*}(\Theta):(W_{e_i},\sigma_i)\in\overline{\Theta}_{\varepsilon},\forall i\in[k_*]\Big\}.
\end{align*}
Subsequently, we demonstrate that $\mathcal{Q}$ is an $\varepsilon$-cover of the metric space $(\mathcal{P}_{k_*}(\Theta),\|\cdot\|_1)$. In other words, we need to show that for any $p_{G}(Y|X)\in\mathcal{P}_{k_*}(\Theta)$, there exists some density $p_{\overline{G}}\in\mathcal{Q}$ such that $\|p_{G}-p_{\overline{G}}\|_1\lesssim\varepsilon$. For this sake, let us consider $P:=\binom{k_*}{K}$ $K$-element subsets of $[k_*]$, which are of the form $\{\tau_1,\ldots,\tau_K\}$ for any $\tau\in[P]$. Additionally, we also denote $\{\tau_{K+1},\ldots,\tau_{k_*}\}:=[k_*]\setminus\{\tau_1,\ldots,\tau_K\}$, and then partition the covariate space $\mathcal{X}$ in two ways as follows:
\begin{align*}
    \mathcal{X}_{\tau}&:=\{x\in\mathcal{X}:|g(x,W_{e_i})|\geq |g(x,W_{e_{j}})|,\forall i\in\{\tau_1,\ldots,\tau_K\},j\in\{\tau_{K+1},\ldots,\tau_{k_*}\}\}\\
    \overline{\mathcal{X}}_{\tau}&:=\{x\in\mathcal{X}:|g(x,\overline{W}_{e_i})|\geq |g(x,\overline{W}_{e_j})|,\forall i\in\{\tau_1,\ldots,\tau_K\},j\in\{\tau_{K+1},\ldots,\tau_{k_*}\}\}.
\end{align*}
As $\Theta$ and $\mathcal{X}$ are bounded, we may assume that $\|W_{e_i}-\overline{W}_{e_i}\|\leq C\eta$ for any parameter $W_e$, where $\eta$ is some positive constant and $C>0$ will be chosen later. Similarly, since $\mathcal{X}$ is a bounded set, then for any $\tau\in[P]$, we can find some constant $c_\tau\geq 0$ that satisfies
\begin{align*}
    \min_{\substack{x\in\mathcal{X}_{\tau}, \ i\in\{\tau_1,\ldots,\tau_K\}, \\ j\in\{\tau_{K+1},\ldots,\tau_{k_*}\}}}[|g(x,W_{e_i})|-|g(x,W_{e_j})|]=c_{\tau}\eta,
\end{align*}
Now, we point out that $\mathcal{X}_{\tau}\subseteq\overline{\mathcal{X}}_{\tau}$ if $c_{\tau}>0$, and $\mathcal{X}_{\tau}$ has measure zero otherwise. 

\textbf{Case 1: $c_{\tau}>0$.}

Recall that $g(x,W_e)$ is Lipschitz continuous with respect to $W_e$, i.e.
\begin{align*}
    |g(x,W_{e_i})-g(x,\overline{W}_{e_i})|\leq L(x)\|W_{e_i}-\overline{W}_{e_i}\|,
\end{align*}
where $L(x)$ is some positive constant depending on $x$. Since $\mathcal{X}$ is bounded, we get that $L(x)\leq A$ for some positive constant, which follows that $|g(x,W_{e_i})-g(X,\overline{W}_{e_i})|\leq AC\eta$. As a result, for $x\in\mathcal{X}_{\tau}$, $i\in\{\tau_1,\ldots,\tau_K\}$ and $j\in\{\tau_{K+1},\ldots,\tau_{k_*}\}$, we observe that
\begin{align*}
    |g(x,\overline{W}_{e_i})|&\geq |g(x,W_{e_i})|-|g(x,W_{e_i})-g(x,\overline{W}_{e_i})|\\
    &\geq |g(x,W_{e_j})|+c_{\tau}\eta-AC\eta\\
    &\geq |g(x,\overline{W}_{e_j})|-|g(x,\overline{W}_{e_j})-g(x,W_{e_j})|+c_{\tau}\eta-AC\eta\\
    &\geq |g(x,\overline{W}_{e_j})|-2AC\eta+c_{\tau}\eta.
\end{align*}
By choosing $C=\frac{c_{\tau}}{2A}$, we obtain that $|g(x,\overline{W}_{e_i})|\geq |g(x,\overline{W}_{e_j})|$ for any $i\in\{\tau_1,\ldots,\tau_K\}$ and $j\in\{\tau_{K+1},\ldots,\tau_{k_*}\}$. Therefore, $x\in\overline{\mathcal{X}}_{\tau}$, and we can conclude that $\mathcal{X}_{\tau}\subseteq\overline{\mathcal{X}}_{\tau}$.

\textbf{Case 2: $c_{\tau}=0$.} 

For $x\in\mathcal{X}_{\tau}$, we assume that
\begin{align*}
    |g(x,W_{e_{\tau_1}})|\geq\ldots\geq|g(x,W_{e_{\tau_K}})|\geq|g(x,W_{e_{\tau_{K+1}}})|\geq\ldots\geq |g(x,W_{e_{\tau_{k_*}}})|.
\end{align*}
Now that $c_{\tau}=0$, we have
\begin{align*}
    |g(x,W_{e_{\tau_K}})|-|g(x,W_{e_{\tau_{K+1}}})|=0.
\end{align*}
As a consequence, we obtain that $\mathcal{X}_{\tau}$ is a subset of 
\begin{align*}
    \mathcal{S}:\{x\in\mathcal{X}:|g(x,W_{e_{\tau_K}})|=|g(x,W_{e_{\tau_{K+1}}})|\}.
\end{align*}
Since $\mathcal{S}$ has measure zero, $\mathcal{X}_{\tau}$ also inherits this property.

Thus, we already proved that $\mathcal{X}_{\tau}\subseteq\overline{\mathcal{X}}_{\tau}$ if $c_{\tau}>0$, and $\mathcal{X}_{\tau}$ has measure zero otherwise.

Next, let us consider $X\in\mathcal{X}_{\tau}$, where $\tau\in[P]$ such that $\{\tau_1,\ldots,\tau_K\}=\{1,\ldots,K\}$, 
we can check that $H_n:=\Big[\sum_{j=1}^{K}\exp(|g(X,\overline{W}_{e_j})|)\Big]\cdot[p_{G}(Y|X)-p_{\overline{G}}(Y|X)]$ can be decomposed as
\begin{align*}
    H_n=&\sum_{i=1}^{K}\exp(|g(X,W_{e_i}|))\Big[f(Y|g(X,W_{e_i}),\sigma_i)-f(Y|g(X,\overline{W}_{e_i}),\overline{\sigma}_i)\Big]\\
    &+\sum_{i=1}^{K}\Big[\exp(|g(X,W_{e_i})|)-\exp(|g(X,\overline{W}_{e_i})|)\Big]\cdot\Big[f(Y|g(X,\overline{W}_{e_i}),\overline{\sigma}_i)-p_{G}(Y|X)\Big].
\end{align*}
As $\Theta$ and $\mathcal{X}$ are bounded, we may assume that $\exp(|g(X,W_{e_i}|))\leq B_1$ and $|f(Y|g(X,\overline{W}_{e_i}),\overline{\sigma}_i)-p_{G}(Y|X)|\leq B_2$ for some positive constants $B_1,B_2$. Thus, we obtain that
\begin{align*}
    |H_n|\lesssim \sum_{i=1}^{K}B_1\cdot\Big[\|W_{e_i}-\overline{W}_{e_i}\|+|\sigma_i-\overline{\sigma}_i|\Big]+\sum_{i=1}^{K}B_2\cdot\|W_{e_i}-\overline{W}_{e_i}\|\lesssim \varepsilon.
\end{align*}
Additionally, since $\Big[\sum_{j=1}^{K}\exp(|g(X,\overline{W}_{e_j})|)\Big]\cdot[p_{G}(Y|X)-p_{\overline{G}}(Y|X)]\geq K$, we also achieve that $|p_{G}(Y|X)-p_{\overline{G}}(Y|X)|\lesssim\varepsilon$. Analogously, we can show that this inequality holds for $X\in\mathcal{X}_{\tau}$ for any $\tau\in[P]$. Since $\bigcup_{p=1}^{P}\mathcal{X}_{\tau}=\mathcal{X}$, we also deduce that the previous bound holds for any $X\in\mathcal{X}$. Consequently, it follows that $\|p_{G}(X,Y)-p_{\overline{G}}(Y|X)\|\lesssim\varepsilon$. This result indicates that $\mathcal{Q}$ is an $\varepsilon$-cover of the metric space $(\mathcal{P}_{k_*}(\Theta),\|\cdot\|_1)$. Therefore, we get
\begin{align*}
    N(\varepsilon,\mathcal{P}_{k_*}(\Theta),\|\cdot\|_1)\leq |\overline{\Theta}_{\varepsilon}|\leq \mathcal{O}(\varepsilon^{-(q+1)k_*}),
\end{align*}
or equivalently,
\begin{align*}
    \log N(\varepsilon,\mathcal{P}_{k_*}(\Theta),\|\cdot\|_1)\leq |\overline{\Theta}_{\varepsilon}|\lesssim \log(1/\varepsilon).
\end{align*}
\textbf{Part (b).} Firstly, we will derive an upper bound for the Gaussian experts $f(Y|g(X,W_e),\sigma)$. Since $\Theta$ is a compact set, we have $|g(X,W_e)|\leq m_1$ and $m_2\leq \sigma\leq m_3$ for any $X\in\mathcal{X}$ and $(W_e,\sigma)\in\Theta$. Then, it follows that $f(Y|g(X,W_e),\sigma)\leq B(X,Y)$, where
\begin{align*}
    B(Y|X)=\begin{cases}
        \dfrac{1}{\sqrt{2\pi}m_2}\exp(-Y^2/(8m_3^2)), \hspace{1cm} \text{for } |Y|\geq 2m_1\\
        \dfrac{1}{\sqrt{2\pi}m_2}, \hspace{3.7cm} \text{for } |Y|<2m_1,
    \end{cases}
\end{align*}
for any $X\in\mathcal{X}$. Next, let $\eta\leq\varepsilon$ be some positive constant that we choose later, then we denote $\{h_1,\ldots,h_N\}$ as an $\eta$-cover over $\mathcal{P}_{k_*}(\Theta)$. Based on this cover, we build the following brackets $L_i(Y|X):=\max\{h_i(Y|X)-\eta,0\}$ and $U_i(Y|X):=\max\{h_i(Y|X)+\eta,B(Y|X)\}$, for any $i\in[N]$. We can validate that $\mathcal{P}_{k_*}(Y|X)\subseteq\bigcup_{i=1}^{N}[L_i(Y|X),U_i(Y|X)]$ and $U_i(X,Y)-L_i(X,Y)\leq\min\{2\eta,B(Y|X)\}$. Moreover, let $M:=\max\{2m_1,\sqrt{8}m_3\}\log(1/\eta)$, we have
\begin{align*}
    &\|U_i-L_i\|_1=\int[U_i(Y|X)-L_i(Y|X)]\dint (X,Y)\\
    &\leq \int_{|Y|<2m_1}[U_i(Y|X)-L_i(Y|X)]\dint(X,Y)+\int_{|Y|\geq 2m_1}[U_i(Y|X)-L_i(Y|X)]\dint (X,Y)\\
    &\leq M\eta+\exp(-M^2/(2m_3^2))\leq c\eta,
\end{align*}
where $c$ is some positive universal constant. The above result implies that
\begin{align*}
    H_B(c\eta,\mathcal{P}_{k_*}(\Theta),\|\cdot\|_1)\leq \log N(\eta,\mathcal{P}_{k_*}(\Theta),\|\cdot\|_1)\lesssim\log(1/\eta).
\end{align*}
Then, we choose $\eta=\varepsilon/c$, we arrive at $H_B(\varepsilon,\mathcal{P}_{k_*}(\Theta),\|\cdot\|_1)\lesssim\log(1/\varepsilon)$. Furthermore, as $\ell_1$-norm is upper bounded by the Hellinger distance $h$, we obtain that
\begin{align*}
    H_B(\varepsilon,\mathcal{P}_{k_*}(\Theta),h)\lesssim\log(1/\varepsilon).
\end{align*}
Hence, the proof is completed.

\subsection{Proof of Theorem~\ref{theorem:parameter_estimation}}
\label{appendix:parameter_estimation}
First of all, we will demonstrate the Hellinger bound given in \eqref{eq:Hellinger_lower_bound}. Since the Hellinger distance is lower bounded by the Total Variation distance, i.e. $h\geq V$, it is sufficient to show that
\begin{align*}
    \mathbb{E}_X[V(p_{G}(\cdot|X),p_{G_*}(\cdot|X))]\gtrsim \mathcal{D}(G,G_*).
\end{align*}
For this purpose, we split the proof of this bound into two parts which we refer to as local part and global part.

\textbf{Local part:} In this part, we will derive the following inequality:
\begin{align}
    \label{eq:local_part}
    \lim_{\varepsilon\to0}\inf_{G\in\mathcal{E}_{k_*}(\Theta):\mathcal{D}(G,G_*)\leq\varepsilon}\mathbb{E}_X[V(p_{G}(\cdot|X),p_{G_*}(\cdot|X))]/\mathcal{D}(G,G_*)>0.
\end{align}
Assume by contrary that this claim does not hold, then there exists a sequence of mixing measures $G_n:=\sum_{i=1}^{k_*}\delta_{(W^n_{e_i},\sigmain)}\in\mathcal{E}_{k_*}(\Theta)$ that satisfies two following limits: $\mathbb{E}_X[V(p_{G_n}(\cdot|X),p_{G_*}(\cdot|X))]/\mathcal{D}(G_n,G_*)\to0$ and $\mathcal{D}(G_n,G_*)\to0$ when $n\to\infty$. Now, we consider Voronoi cells $\mathcal{A}^n_j:=\mathcal{A}_j(G_n)$ defined as in \eqref{eq:Voronoi_cells}. As we use asymptotic arguments in this proof, we may assume without loss of generality (WLOG) that the above Voronoi cells do not depend on the sample size $n$, that is, $\mathcal{A}^n_j=\mathcal{A}_j$. Additionally, since each Vornoi cell has only one element, we can also assume that $\mathcal{A}_j=\{j\}$ for any $j\in[k_*]$. This together with the fact that $\mathcal{D}(G_n,G_*)\to0$ indicates that $(W^n_{e_j},\sigma^n_j)\to (W^*_{e_j},\sigma^*_j)$ as $n\to\infty$ for all $j\in[k_*]$. Subsequently, we assume WLOG that
\begin{align}
    \label{eq:Voronoi_loss_assume}
    \mathcal{D}(G_n,G_*):=\sum_{j=1}^{K}\Big[\|W^n_{e_j}-W^*_{e_j}\|+|\sigma^n_j-\sigma^*_j|\Big].
\end{align}
Next, we will partition the covariate space $\mathcal{X}$ in order to specify the value of top-K sparse softmax gating function in the conditional density $p_{G_*}(Y|X)$. In particular, we consider $P:=\binom{k_*}{K}$ different subsets of $\{1,2,\ldots,k_*\}$, each of which contains $K$ elements. Moreover, we denote these subsets as $\{\tau_1,\tau_2,\ldots,\tau_K\}$, and let $\{\tau_{K+1},\ldots,\tau_{k_*}\}:=[k_*]\setminus\{\tau_1,\tau_2,\ldots,\tau_K\}$ for any $\tau\in[P]$.  Then, we divide the covariate space $\mathcal{X}$ into smaller regions as follows:
\begin{align*}
    \mathcal{X}_{*,\tau}&:=\{x\in\mathcal{X}:|g(x,W^*_{e_i})|\geq |g(x,W^*_{e_j})|, \forall i\in\{\tau_1,\ldots,\tau_K\},j\in\{\tau_{K+1},\ldots,\tau_{k_*}\}\},\\
    \mathcal{X}_{n,\tau}&:=\{x\in\mathcal{X}:|g(x,W^n_{e_i})|\geq |g(x,W^n_{e_j})|, \forall i\in\{\tau_1,\ldots,\tau_K\},j\in\{\tau_{K+1},\ldots,\tau_{k_*}\}\}.
\end{align*}
Now, we show that $\mathcal{X}_{*,\tau}\subseteq\mathcal{X}_{n,\tau}$ for sufficiently large $n$ for any $\tau\in[P]$. Given some $\eta>0$, since $W^n_{e_j}\to W^*_{e_j}$ for any $j\in[k_*]$, we have that $\|W^n_{e_j}-W^*_{e_j}\|\leq C\eta$ for sufficiently large $n$, where $C>0$ will be chosen later. Moreover, as $\mathcal{X}$ is a bounded set, there exists some positive constant $c_{\tau}$ such that
\begin{align*}
    \min_{\substack{x\in\mathcal{X}_{\tau}, \ i\in\{\tau_1,\ldots,\tau_K\}, \\ j\in\{\tau_{K+1},\ldots,\tau_{k_*}\}}}[|g(x,W^*_{e_i})|-|g(x,W^*_{e_j})|]=c_{\tau}\eta.
\end{align*}
By arguing similarly to Case 2 in part (a) of Appendix~\ref{appendix:upper_bounds}, we get that if $c_{\tau}=0$, then $\mathcal{X}_{*,\tau}$ has measure zero. Therefore, we may assume that $c_{\tau}>0$. Let $x\in\mathcal{X}_{*,\tau}$ and follow the same arguments used for Case 1 in part (a) of Appendix~\ref{appendix:upper_bounds}, we obtain that 
\begin{align*}
    |g(x,W^n_{e_i})|\geq |g(x,W^n_{e_j})| - 2AC\eta+c_{\tau}\eta,
\end{align*}
for sufficiently large $n$, $i\in\{\tau_1,\ldots,\tau_K\}$ and $j\in\{\tau_{K+1},\ldots,\tau_{k_*}\}$, where $A$ is an upper bound of the Lipschitz constant of expert function $g(x,W_e)$. Then, by setting $C=\frac{c_{\tau}}{2A}$, we achieve that $|g(x,W^n_{e_i})|\geq|g(x,W^n_{e_j})|$ for sufficiently large $n$, $i\in\{\tau_1,\ldots,\tau_K\}$ and $j\in\{\tau_{K+1},\ldots,\tau_{k_*}\}$. This result indicates that $x\in\mathcal{X}_{n,\tau}$, and thus, $\mathcal{X}_{*,\tau}\subseteq\mathcal{X}_{n,\tau}$ for sufficiently large $n$.

For $(X,Y)\in\mathcal{X}_{*,\tau}\times\mathbb{R}$ where $\tau\in[P]$ such that $\{\tau_1,\ldots,\tau_K\}=\{1,\ldots,K\}$, two conditional densities $p_{G_n}(Y|X)$ and $p_{G_*}(Y|X)$ are represented as
\begin{align*}
    p_{G_*}(Y|X)&=\sum_{i=1}^{K}\dfrac{\exp(|g(X,W^*_{e_i})|)}{\sum_{j=1}^{K}\exp(|g(X,W^*_{e_j})|)}\cdot f(Y|g(X,W^*_{e_i}),\sigma^*_i),\\
    p_{G_n}(Y|X)&=\sum_{i=1}^{K}\dfrac{\exp(|g(X,W^n_{e_i}))|}{\sum_{j=1}^{K}\exp(|g(X,W^n_{e_j})|)}\cdot f(Y|g(X,W^n_{e_i}),\sigma^n_i).
\end{align*}
Then, it can be verified that the quantity $Z_n:=\Big[\sum_{j=1}^{K}\exp(|g(X,W^*_{e_j})|)\Big]\cdot [p_{G_n}(Y|X)-p_{G_*}(Y|X)]$ can be decomposed as
\begin{align*}
    Z_n&=\sum_{i=1}^{K}\Big[\exp(|g(X,W^n_{e_i})|)f(Y|g(X,W^n_{e_i}),\sigma^n_i)-\exp(|g(X,W^*_{e_i})|)f(Y|g(X,W^*_{e_i}),\sigma^*_i)\Big]\\
    &\quad -\sum_{i=1}^{K}\Big[\exp(|g(X,W^n_{e_i})|)-\exp(|g(X,W^*_{e_i})|)]p_{G_n}(Y|X).
\end{align*}
Our upcoming arguments are separated into three steps as follows:

\textbf{Step 1: Taylor expansion.} In this step, we apply the first-order Taylor expansions to the first and second terms in the decomposition of $Z_n$ as
\begin{align*}
    Z_n&=\sum_{i=1}^{K}\Bigg\{\sum_{u=1}^{q}(\Delta W^n_{e_i})^{(u)}\frac{\partial g}{\partial W^{(u)}_{e}}(X,W^*_{e_i})\exp(|g(X,W^*_{e_i})|)\Big[s^*_{g,i}(X)\cdot f(Y|g(X,W^*_{e_i}),\sigma^*_i)\\
    &+\frac{\partial f}{\partial g}(Y|g(X,W^*_{e_i}),\sigma^*_i)\Big]+\frac{1}{2}(\Delta\sigma^n_i)\exp(|g(X,W^*_{e_i})|)\frac{\partial^2 f}{\partial g^2}(Y|g(X,W^*_{e_i}),\sigma^*_i)\Bigg\} +R_1(X,Y)\\
    &\quad -\sum_{i=1}^{K}\sum_{u=1}^{q}(\Delta W^n_{e_i})^{(u)}\frac{\partial g}{\partial W^{(u)}_{e}}(X,W^*_{e_i})\exp(|g(X,W^*_{e_i})|)s^*_{g,i}(X)\cdot p_{G_n}(Y|X) + R_2(X,Y).
\end{align*}
Here, we denote $\Delta W^n_{e_i}:=W^n_{e_i}-W^*_{e_i}$, $\Delta\sigma^n_i:=\sigma^n_i-\sigma^*_i$ and $s^*_{g,i}(X):=sign(g(X,W^*_{e_i}))$. Meanwhile, $R_j(X,Y)$ is a Taylor remainder such that $R_j(X,Y)/\mathcal{D}(G_n,G_*)\to0$ as $n\to\infty$ for $j\in\{1,2\}$. From the above equation, $Z_n-R_1(X,Y)-R_2(X,Y)$ can be seen as a linear combination of elements of the set $\mathcal{F}:=\bigcup_{i=1}^{K}\bigcup_{\tau=0}^{3}\mathcal{F}_{\tau,i}$ where we define
\begin{align*}
    \mathcal{F}_{0,i}&:=\left\{\frac{\partial g}{\partial W^{(u)}_{e}}(X,W^*_{e_i})\exp(|g(X,W^*_{e_i})|)s^*_{g,i}(X)\cdot f(Y|g(X,W^*_{e_i}),\sigma^*_i):1\leq u\leq q\right\},\\
    \mathcal{F}_{1,i}&:=\left\{\frac{\partial g}{\partial W^{(u)}_{e}}(X,W^*_{e_i})\exp(|g(X,W^*_{e_i})|)\frac{\partial f}{\partial g}(Y|g(X,W^*_{e_i}),\sigma^*_i):1\leq u\leq q\right\},\\
    \mathcal{F}_{2,i}&:=\left\{\exp(|g(X,W^*_{e_i})|)\frac{\partial^2 f}{\partial g^2}(Y|g(X,W^*_{e_i}),\sigma^*_i)\right\},\\
    \mathcal{F}_{3,i}&:=\left\{\frac{\partial g}{\partial W^{(u)}_{e}}(X,W^*_{e_i})\exp(|g(X,W^*_{e_i})|)s^*_{g,i}(X)\cdot p_{G_n}(Y|X):1\leq u\leq q\right\}.
\end{align*}
\textbf{Step 2: Non-vanishing coefficients.} Let $E_{0,i}(u)$, $E_{1,i}(u)$, $E_{2,i}$ and $E_{3,i}(u)$ be coefficients associated with those elements in the aforementioned linear combination. Then, we have $E_{0,i}(u)=E_{1,i}(u)=-E_{3,i}(u)=(\Delta W^n_{e_i})^{(u)}$ and $E_{2,i}=(\Delta\sigma^n_i)/2$ for any $i\in[K]$ and $u\in[q]$. We will show that at least one among $E_{0,i}(u)/\mathcal{D}(G_n,G_*)$, $E_{1,i}(u)/\mathcal{D}(G_n,G_*)$, $E_{2,i}/\mathcal{D}(G_n,G_*)$ and $E_{3,i}(u)/\mathcal{D}(G_n,G_*)$ does not approach zero when $n$ goes to infinity. Assume by contrary that all of them vanish as $n\to\infty$. Thus, it follows that
\begin{align*}
    &\frac{\|\Delta W^n_{e_i}\|_1}{\mathcal{D}(G_n,G_*)}=\sum_{u=1}^{q}\frac{|E_{0,i}(u)|}{\mathcal{D}(G_n,G_*)}\to0,\\
    &\frac{|\Delta\sigma^n_i|}{\mathcal{D}(G_n,G_*)}=\frac{2|E_{2,i}|}{\mathcal{D}(G_n,G_*)}\to0.
\end{align*}
Putting these limits and the formulation of $\mathcal{D}(G_n,G_*)$ in \eqref{eq:Voronoi_loss_assume} together, we obtain that
\begin{align*}
    1=\frac{\mathcal{D}(G_n,G_*)}{\mathcal{D}(G_n,G_*)}=\sum_{i=1}^{K}\frac{\|\Delta W^n_{e_i}\|_1}{\mathcal{D}(G_n,G_*)}+\sum_{i=1}^{K}\frac{|\Delta\sigma^n_i|}{\mathcal{D}(G_n,G_*)}\to0,
\end{align*}
as $n\to\infty$, which is a contradiction. As a result, not all the ratios $E_{0,i}(u)/\mathcal{D}(G_n,G_*)$, $E_{1,i}(u)/\mathcal{D}(G_n,G_*)$, $E_{2,i}/\mathcal{D}(G_n,G_*)$ and $E_{3,i}(u)/\mathcal{D}(G_n,G_*)$ vanish when $n$ goes to infinity.

\textbf{Step 3: Fatou's arguments.} By applying the Fatou's lemma, we have
\begin{align*}
    0=\lim_{n\to\infty}\dfrac{\mathbb{E}_X[V(p_{G}(\cdot|X),p_{G_*}(\cdot|X))]}{\mathcal{D}(G_n,G_*)}=\frac{1}{2}\int\liminf_{n\to\infty}\dfrac{|p_{G_n}(Y|X)-p_{G_*}(Y|X)|}{\mathcal{D}(G_n,G_*)}\dint X\dint Y,
\end{align*}
which implies that $[p_{G_n}(Y|X)-p_{G_*}(Y|X)]/\mathcal{D}(G_n,G_*)\to0$ as $n\to\infty$ for almost surely $(X,Y)$. As a result, this limit also holds for almost surely $(X,Y)\in\mathcal{X}_{*,\tau}\times\mathbb{R}$ where $\tau\in[P]$ such that $\{\tau_1,p_2,\ldots,\tau_K\}:=\{1,2,\ldots,K\}$. Let us denote by $m_n$ the maximum of the absolute values of these four ratios. Since at least one among them does not tend to zero as $n\to\infty$, we have $1/m_n\not\to\infty$. Additionally, we denote
\begin{align*}
   \frac{(\Delta W^n_{e_i})^{(u)}}{m_n\mathcal{D}(G_n,G_*)}\to\alpha_{i}(u), \qquad \frac{(\Delta\sigma^n_i)/2}{m_n\mathcal{D}(G_n,G_*)}\to\beta_{i},
\end{align*}
as $n\to\infty$ for any $i\in[K]$ and $u\in[q]$ with a note that at least one among $\alpha_{u,i},\beta_i$ is non-zero. Then, it follows from the fact $[p_{G_n}(Y|X)-p_{G_*}(Y|X)]/\mathcal{D}(G_n,G_*)\to$ as $n\to\infty$ that
\begin{align}
    0&=\lim_{n\to\infty}\dfrac{\Big[\sum_{j=1}^{K}\exp(|g(X,W^*_{e_j})|)\Big]\cdot\Big[p_{G_n}(Y|X)-p_{G_*}(Y|X)\Big]}{m_n\mathcal{D}(G_n,G_*)}\nonumber\\
    &=\lim_{n\to\infty}\dfrac{Z_n}{m_n\mathcal{D}(G_n,G_*)}\nonumber\\
    \label{eq:limit}
    &=\sum_{i=1}^{K}\Big[\sum_{\tau=0}^{2}T_{\tau,i}(X)s^*_{g,i}(X)\cdot\frac{\partial^{\tau}f}{\partial g^{\tau}}(Y|g(X,W^*_{e_i}),\sigma^*_i)+T_{3,i}(X)s^*_{g,i}(X)\cdot p_{G_*}(Y|X)\Big],
\end{align}
for almost surely $(X,Y)\in\mathcal{X}_{*,\tau}\times\mathbb{R}$, where we define
\begin{align*}
    &T_{0,i}(X)=T_{1,i}(X)=-T_{3,i}(X):=\sum_{u=1}^{q}\alpha_{u,i}\frac{\partial g}{\partial W^{(u)}_e}(X,W^*_{e_i})\exp(|g(X,W^*_{e_i})|),\\
    &T_{2,i}(X):=\beta_i\exp(|g(X,W^*_{e_i})|).
\end{align*}
Note that for almost surely $X\in\mathcal{X}_{*,\tau}$, the set
\begin{align*}
    \left\{s^*_{g,i}(X)\cdot \frac{\partial^{\tau}f}{\partial g^{\tau}}(Y|g(X,W^*_{e_i}), \ s^*_{g,i}(X)\cdot p_{G_*}(Y|X):0\leq\tau\leq 2\right\}
\end{align*}
is linearly independent w.r.t $Y$. Thus, the result in \eqref{eq:limit} implies that $T_{\tau,i}(X)=0$ for almost surely $X\in\mathcal{X}_{*,\tau}$ for $0\leq\tau\leq 3$ and $i\in[K]$. In other words, we obtain for any $i\in[K]$ that $\beta_i=0$ and
\begin{align*}
    \sum_{u=1}^{q}\alpha_{u,i}\frac{\partial g}{\partial W^{(u)}_e}(X,W^*_{e_i})=0,
\end{align*}
for almost surely $X\in\mathcal{X}_{*,\tau}$. Due to the assumption on the expert function $g$, the above equation leads to $\alpha_{u,i}=0$ for any $i\in[K]$ and $u\in[q]$. This contradicts the fact that not all the terms $\alpha_{u,i},\beta_i$ are equal to zero. Hence, we reach the conclusion in \eqref{eq:local_part}. 

Consequently, there exists some positive constant $\varepsilon'>0$ that satisfies
\begin{align*}
    \inf_{G\in\mathcal{E}_{k_*}(\Theta):\mathcal{D}(G,G_*)\leq\varepsilon'}\mathbb{E}_X[V(p_{G}(\cdot|X),p_{G_*}(\cdot|X))]/\mathcal{D}(G,G_*)>0
\end{align*}
\textbf{Global part.} Given the above result, it is sufficient to point out that
\begin{align}
    \label{eq:global_part}
    \inf_{G\in\mathcal{E}_{k_*}(\Theta):\mathcal{D}(G,G_*)>\varepsilon'}\mathbb{E}_X[V(p_{G}(\cdot|X),p_{G_*}(\cdot|X))]/\mathcal{D}(G,G_*)>0.
\end{align}
Assume that this claim does not hold true, then we can find a sequence $\widetilde{G}_n\in\mathcal{E}_{k_*}(\Theta)$ such that $\mathcal{D}(\widetilde{G}_n,G_*)>\varepsilon'$ and $\mathbb{E}_X[V(p_{\widetilde{G}_n}(\cdot|X),p_{G_*}(\cdot|X))]\to0$ as $n\to\infty$. Since $\Theta$ is a compact set, we are able to replace $\widetilde{G}_n$ with its subsequence which converges to some mixing measure $\widetilde{G}\in\mathcal{E}_{k_*}(\Theta)$. Recall that $\mathcal{D}(\widetilde{G}_n,G_*)>\varepsilon'$, then we also get that $\mathcal{D}(\widetilde{G},G_*)>\varepsilon'$.

On the other hand, by means of the Fatou's lemma, we have
\begin{align*}
    0=\lim_{n\to\infty}\mathbb{E}_X[2V(p_{\widetilde{G}_n}(\cdot|X),p_{G_*}(\cdot|X))]\geq\int\liminf_{n\to\infty}|p_{\widetilde{G}_n}(Y|X)-p_{G_*}(Y|X)|\dint X\dint Y,
\end{align*}
which follows that $p_{\widetilde{G}}(Y|X)-p_{G_*}(Y|X)=0$ for almost surely $(X,Y)$. Thus, we achieve that $\widetilde{G}\equiv G_*$, or equivalently $\mathcal{D}(\widetilde{G},G_*)=0$. This contradicts to the fact that $\mathcal{D}(\widetilde{G},G_*)>\varepsilon'>0$. 

Hence, we reach the conclusion in \eqref{eq:global_part}, and the proof is completed.

\section{Additional implementation details} \label{app:add_exp}



This section provides the details parameters of our experiments in Section \ref{sec:exp}. 

\subsection{General Setting} \label{app:setting}
The experiments are developed based on the publicly available Sandwich Transformers \citep{press-etal-2020-improving} implementation\footnote{\url{https://github.com/ofirpress/sandwich_transformer}}. However, the pre-training models were conducted on a single A100 GPU. Please note that parallel training on multiple GPUs might yield different results.

\subsection{Pre-training Experiments}
Tab. \ref{tab:A1} provides the detail configurations for pre-training our Switch Transformer~\citep{fedus_switch_2022} and GLaM~\cite{du_glam_2022} with two versions (\textit{small} \& \textit{medium}) on \texttt{Enwik8}, \texttt{Text8} and \texttt{WikiText-103}. 


\begin{table}[!ht]
\centering
\setlength\tabcolsep{3.06pt}

\begin{tabular}{lccccc}
\midrule
Dataset   & Input length & Batch size & Optimizer & Lr   & \# Iterations  (\textit{Small} / Medium) \\ \midrule
\texttt{Enwik8}      & 512          & 48          & Adam      & 7.0e-4 & \textit{60k} / 80k        \\
\texttt{Text}      & 512          & 48          & Adam      & 7.0e-4 & \textit{60k} / 80k         \\
\texttt{WikiText-103} & 512          & 48         & Adam      & 7.0e-4 & \textit{60k} / 80k         \\ \midrule
\end{tabular}
\caption{Detail configurations for pre-training experiments on \texttt{Enwik8}, \texttt{Text8} and \texttt{WikiText-103} datasets. }
\label{tab:A1}
\end{table}

\subsection{Finetuning Experiments}
\noindent For finetuning experiments, we use the same model architecture as in pre-training. Tab. \ref{tab:A2} shows the detail configurations used for finetuning experiments on \texttt{SST-2}, \texttt{SST-5}, \texttt{IMDB}, and \texttt{BANKING77} datasets. 

\begin{table}[!ht]
\centering

\setlength\tabcolsep{4.86pt}
\begin{tabular}{lccccc}
\midrule
Dataset   & Input length & Batch size & Optimizer & Lr   & \# Epochs \\ \midrule
\texttt{SST-2}     & 512          & 16         & Adam      & 1e-4 & 5         \\
\texttt{SST-5}     & 512          & 16         & Adam      & 1e-4 & 5         \\
\texttt{IMDB}      & 512          & 4          & Adam      & 1e-4 & 5         \\
\texttt{BANKING77} & 512          & 16         & Adam      & 1e-4 & 50         \\ \midrule
\end{tabular}
\caption{Detail configurations for finetuning experiments on four different datasets. }
\label{tab:A2}
\end{table}

\subsection{Significant Results}
We investigate the significant of our pre-training experiments in Section~\ref{sec:pretrain}. We consider the tiny transformer configuration and repeat the training process five times using SMoE and CompeteSMoE. Table~\ref{tab:A5} reports the result of each run, the average and standard devation values. We can see that CompeteSMoE achieves better results than SMoE in all runs, resulting better average and also smaller standard deviations. The two-sided t-test returns the $p-$value at 0.016, suggesting the improvements of CompeteSMoE is significant.

Furthermore, we also note that the training overhead of CompeteSMoE is also quite marginal, only at around $10\%$ longer training time. Overall, the experimental results show that CompeteSMoE provides better pre-training and transfer learning capabilities with low training overheads.
\begin{table}[!ht]
\centering

\setlength\tabcolsep{4.86pt}
\begin{tabular}{lcc}
\toprule
Enwik8 (BPC)   & CompeteSMoE & SMoE  \\ \midrule
\texttt{1st}     & \textbf{1.303}          & 1.333          \\
\texttt{2nd}     & \textbf{1.303}          & 1.322          \\
\texttt{3rd}     & \textbf{1.307}          & 1.315           \\
\texttt{4th}     & \textbf{1.315}          & 1.320           \\
\texttt{5th}     & \textbf{1.304}          & 1.310           \\ \midrule
\texttt{Average}     & \textbf{1.306}          & 1.320        \\
\texttt{}     & $\pm$ 0.005      &  $\pm$ 0.009   \\ \midrule
\texttt{Training epoch time(h)}     & 0.13          & 0.11     \\ 
\bottomrule
\end{tabular}
\caption{BPC of multiple runs on the enwik8 test set using the tiny configuration. }
\label{tab:A5}
\end{table}


\subsection{Routing Visualization} \label{app:visualization}
\label{sec:entropy}
\begin{table}[!ht]
\centering

\setlength\tabcolsep{7.88pt}
\begin{tabular}{lccccccc}
\midrule
Method   & Router 1 & Router 2 & Router 3 & Router 4 & Router 5 & Router 6 & Average \\ \midrule
CompeteSMoE              & 0.3441 & \textbf{0.4015} & 0.5946 & \textbf{0.6810} & \textbf{0.6081} & \textbf{1.3636} & \textbf{0.6655}  \\
                           & $\pm$ 0.29 & $\pm$ 0.37 & $\pm$ 0.39 & $\pm$ 0.38 & $\pm$ 0.37 & $\pm$ 0.35 & $\pm$ 0.36 \\
SMoE      & \textbf{0.1158} & 0.6411 & \textbf{0.4247} & 0.8331 & 0.9701 & 1.3681 & 0.7255 \\
                           & $\pm$ 0.23 & $\pm$ 0.48 & $\pm$ 0.35 & $\pm$ 0.48 & $\pm$ 0.48 & $\pm$ 0.42 & $\pm$ 0.41\\
SMoE-Fixed      & 2.2473 & 2.5103 & 2.2579 & 2.3565 & 2.3181 & 2.3947 & 2.3942 \\
                           & $\pm$ 0.34 & $\pm$ 0.13 & $\pm$ 0.13 & $\pm$ 0.26 & $\pm$ 0.23 & $\pm$ 0.22 & $\pm$ 0.22 \\
XMoE      & 0.7618 & 0.6792 & 1.1808 & 0.9822 & 0.8439 & 1.4097 & 0.9763 \\
                           & $\pm$ 0.47 & $\pm$ 0.42 & $\pm$ 0.38 & $\pm$ 0.45 & $\pm$ 0.42 & $\pm$ 0.40 & $\pm$ 0.43 \\
StableMoE       & 0.7657 & 1.2659 & 1.1378 & 1.2000  & 1.4161 & 1.8161 & 1.2669 \\ 
                           & $\pm$ 0.71 & $\pm$ 0.58 & $\pm$ 0.56 & $\pm$  0.51 & $\pm$ 0.48 & $\pm$ 0.41 & $\pm$ 0.54 \\\midrule
\end{tabular}
\caption{Average entropy of the router distribution on the \texttt{enwik8} dataset. Lower is better.} \label{tab:entropy}
\end{table}

This experiment investigate the quality of routers learned from various training algorithm. To this end, we consider reporting the entropy from the router's softmax output. Routers with high entropy produce close to uniform policies, indicating a less-confident assignment.  In contrast, low entropy routers are more confident in assigning experts to tokens. Although entropy alone cannot determine the router quality, when combined with the evaluation outcome, one can gauge the learned routers quality. Particularly, if a router has low entropy and achieve better evaluation results, one can expect that the learned routing policy has high quality.

Table~\ref{tab:entropy} reports the entropy of each router in a small model on the enwik8 test set. We can see that CompeteSMoE achieved lower entropy in four out of six routers, and lower average entropy. Together with the better performance, we can conclude that the routers learned by CompeteSMoE have higher quality (better token-expert assignment) than other baselines.

We further visualize the softmax output of the routers from one randomly chosen sample in Figure~\ref{fig:distributions}. In most cases, we can observe the distributional outputs are much sharper in CompeteSMoE (lower entropy) than other baselines.

\begin{figure*}

 \centering
    \begin{subfigure}{\textwidth}
        \includegraphics[width=\linewidth]{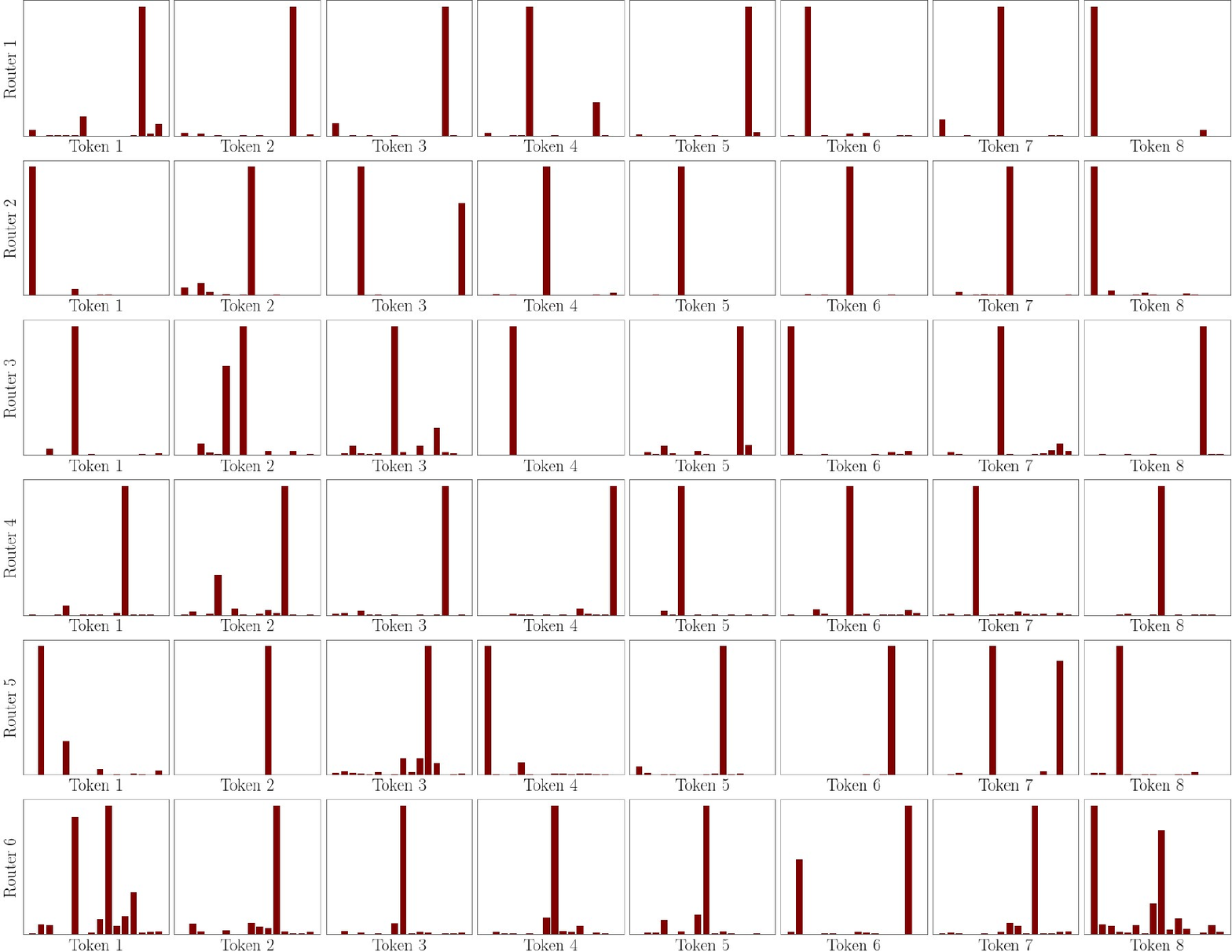}
        \subcaption{CompeteSMoE}
        \label{fig:arm5}
    \end{subfigure}
\medskip
    \begin{subfigure}{\textwidth}
        \includegraphics[width=\linewidth, height=0.4\textheight]{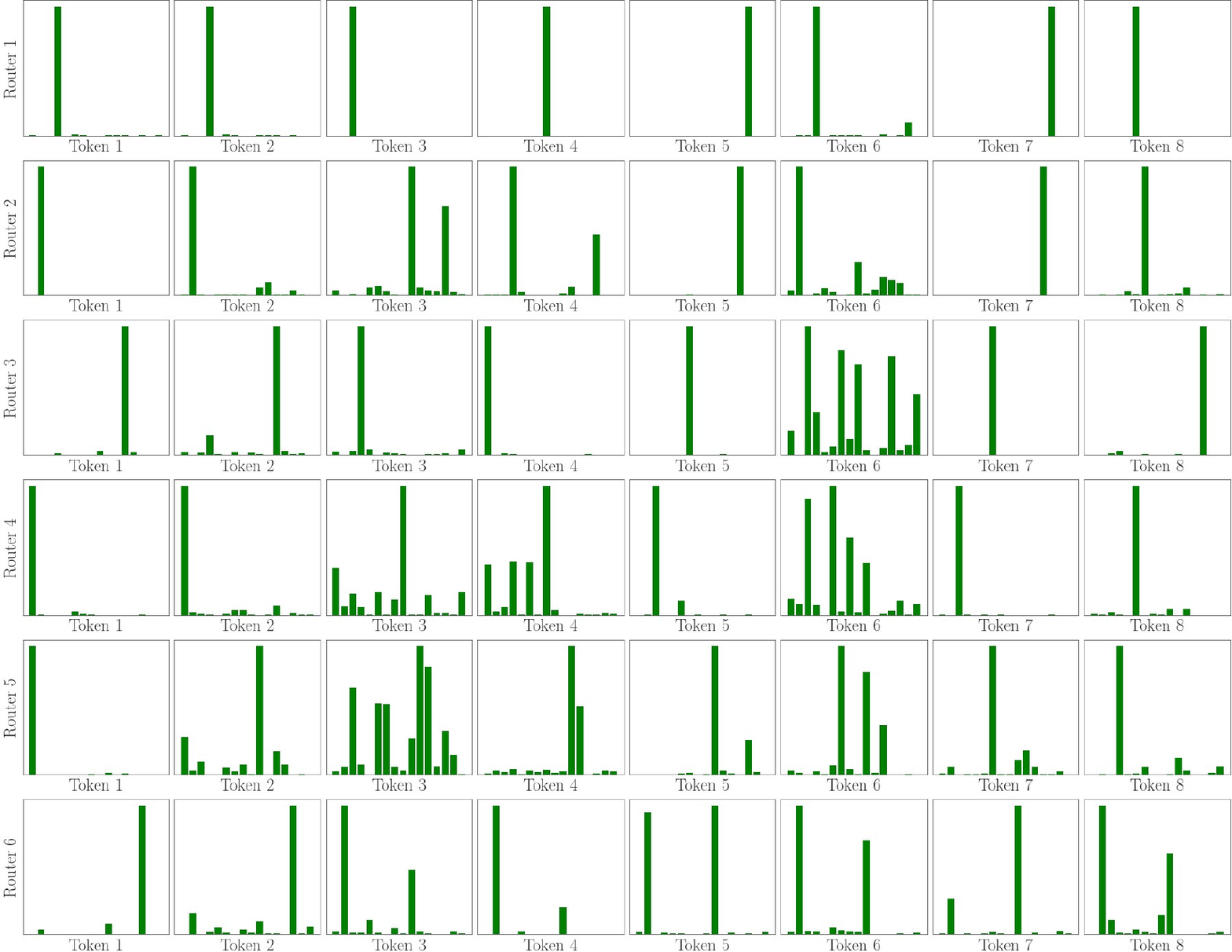}
        \subcaption{ SMoE}
        \label{fig:arm4}
    \end{subfigure}
    \caption{Visualization of the distribution for the output of routers.}%
    \label{fig:distributions}
\end{figure*}%
\begin{figure*}[ht]\ContinuedFloat

 \centering
    \begin{subfigure}{\textwidth}
        \includegraphics[width=\linewidth, height=0.4\textheight]{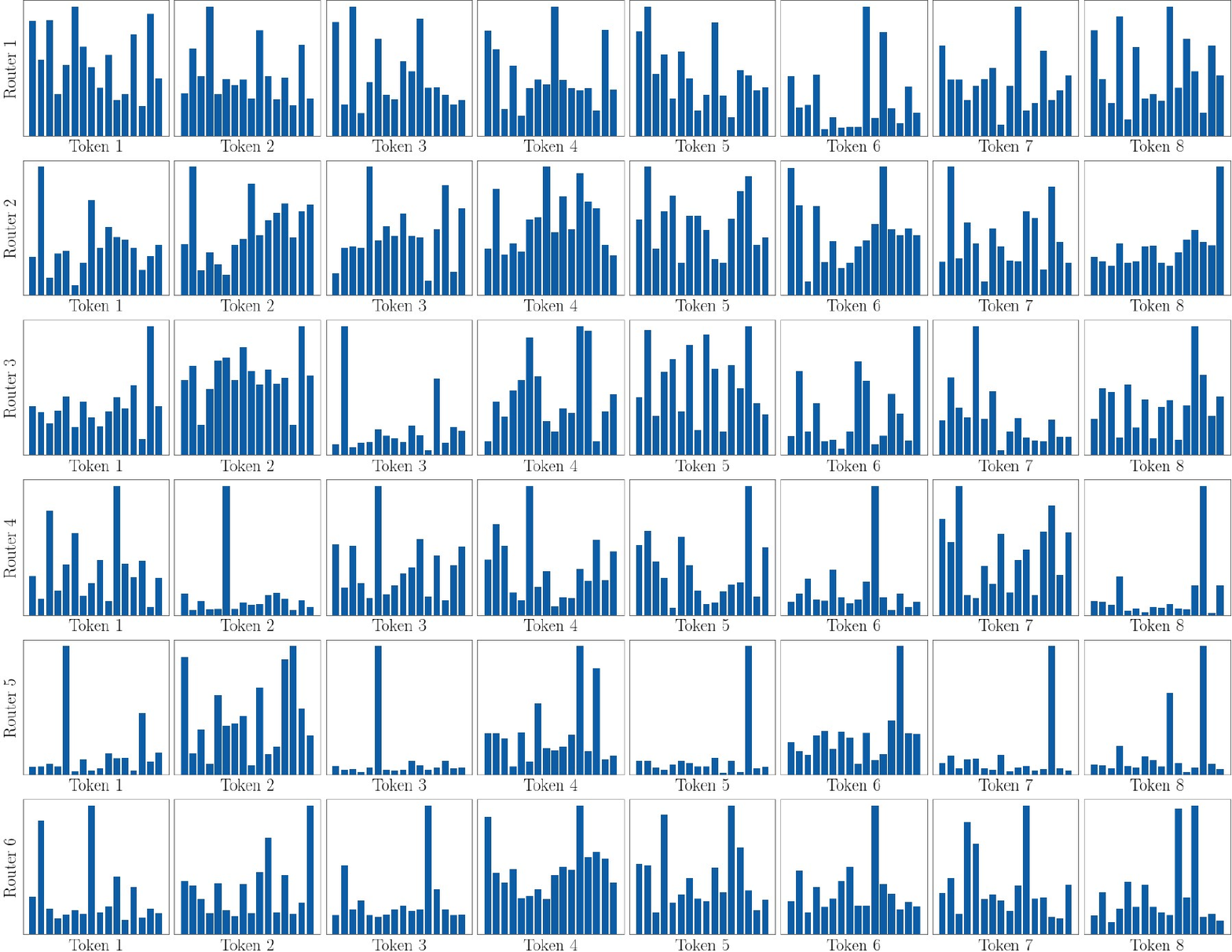}
        \subcaption{ SMoE-Fixed}
        \label{fig:arm1}
    \end{subfigure}
\medskip
  \begin{subfigure}{\textwidth}
        \includegraphics[width=\linewidth, height=0.4\textheight]{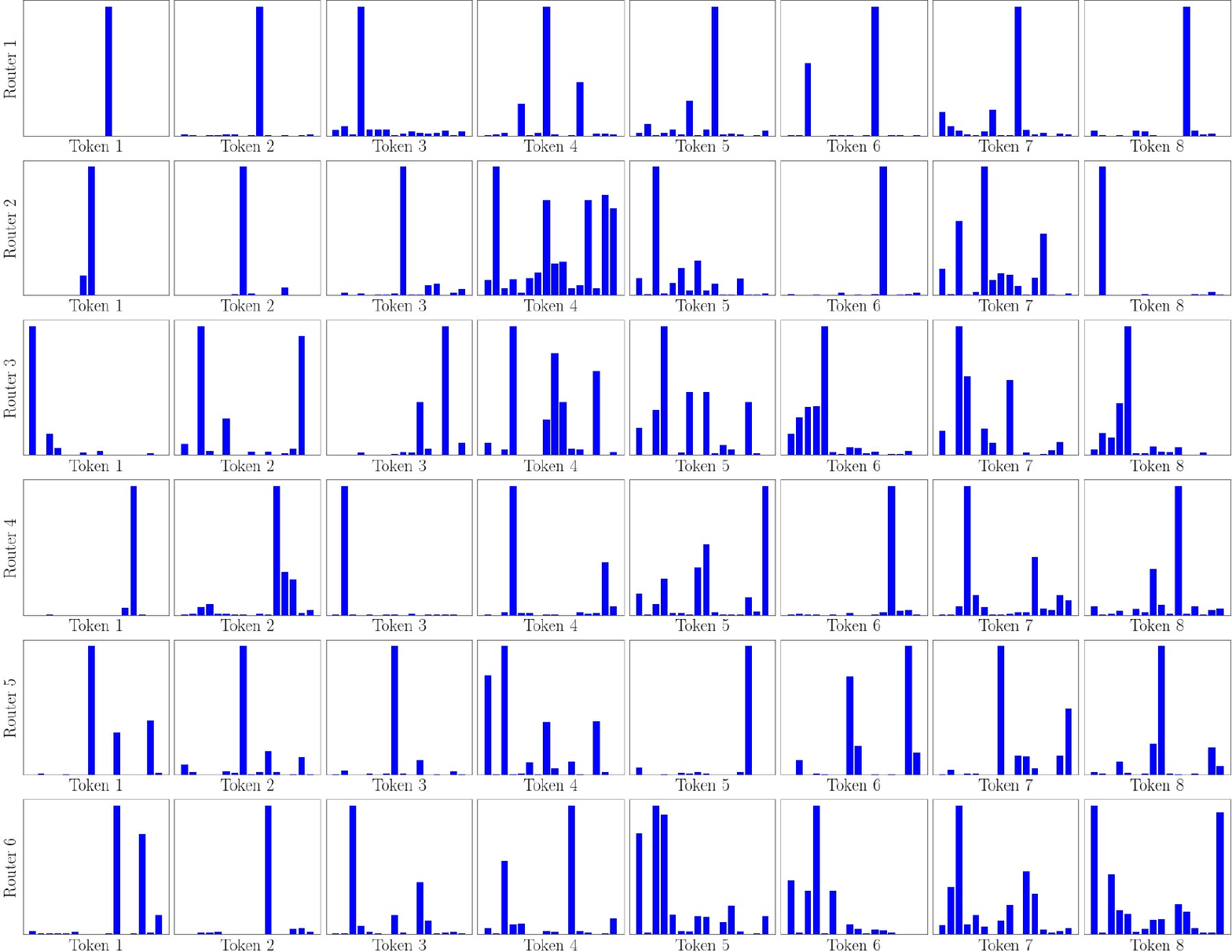}
        \subcaption{ XMoE}
        \label{fig:arm2}
    \end{subfigure}
    \caption{Visualization of the distribution for the output of routers.}

\end{figure*}
\begin{figure*}[ht]\ContinuedFloat
\centering
    \begin{subfigure}{\textwidth}
        \includegraphics[width=\linewidth, height=0.4\textheight]{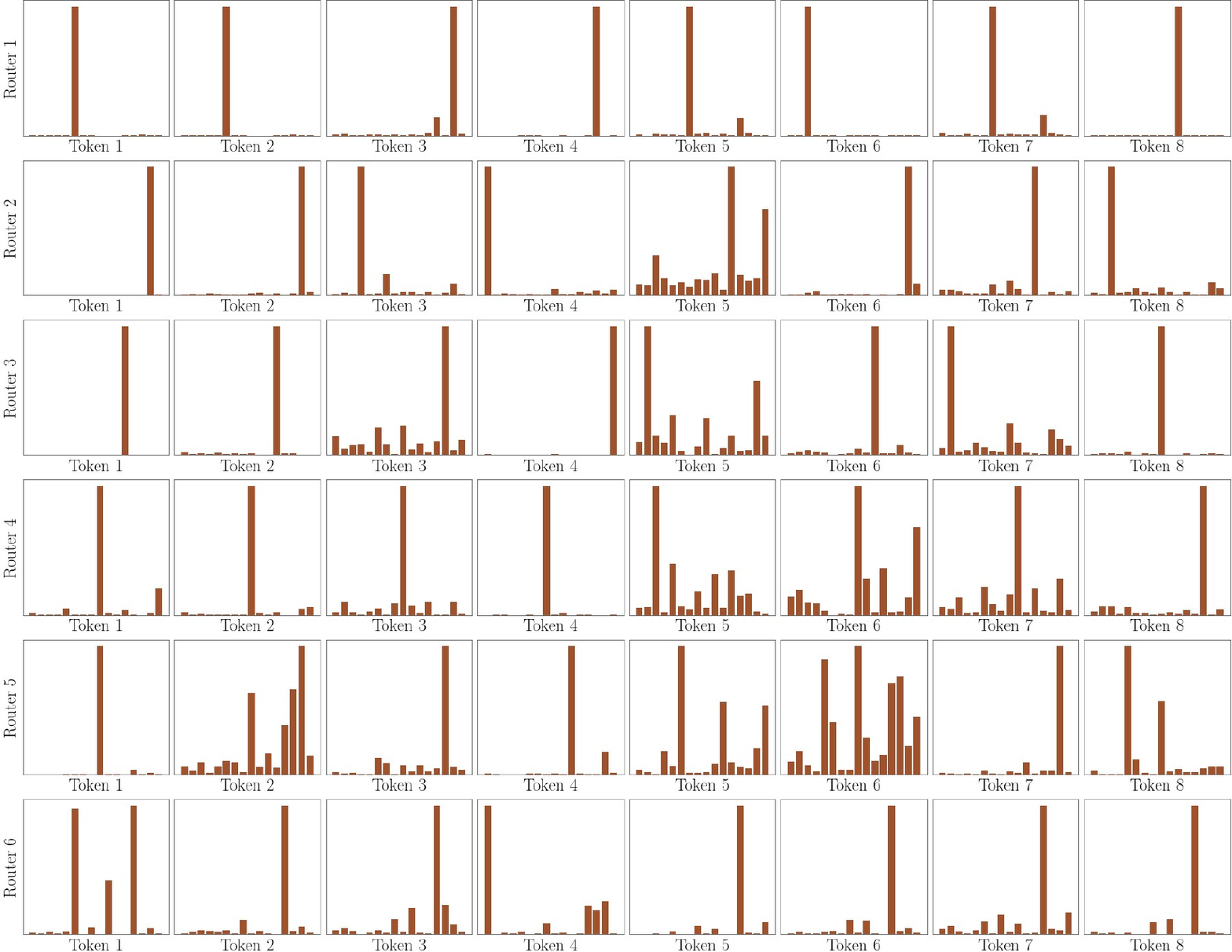}
        \subcaption{ StableMoE}
        \label{fig:arm3}
    \end{subfigure}
    \caption{Visualization of the distribution for the output of routers.}

\end{figure*}

\end{document}

%% file: preambleicml2024.tex
\usepackage{amsmath,amsfonts,bm}

\usepackage{amsthm}
\usepackage{amsmath}
\usepackage{amssymb,bbm}
\usepackage{textcomp}
\usepackage{siunitx}
\usepackage{wrapfig}
\usepackage{multirow}
\usepackage{multicol}
\usepackage{epsf}
\usepackage{epsfig}
\usepackage{fancyhdr}
\usepackage{booktabs}       
\usepackage{amsfonts}       
\usepackage{nicefrac}       
\usepackage{microtype}
\usepackage{soul}
\usepackage{mathtools}

\DeclarePairedDelimiterX{\infdivx}[2]{(}{)}{%
  #1\;\delimsize\|\;#2%
}

\def\eqref#1{equation~\ref{#1}}

\def\1{\bm{1}}

\def\vh{{\bm{h}}}

\def\vs{{\bm{s}}}

\def\vx{{\bm{x}}}
\def\vy{{\bm{y}}}
\def\vz{{\bm{z}}}

\DeclareMathAlphabet{\mathsfit}{\encodingdefault}{\sfdefault}{m}{sl}
\SetMathAlphabet{\mathsfit}{bold}{\encodingdefault}{\sfdefault}{bx}{n}

\def\gE{{\mathcal{E}}}

\def\gL{{\mathcal{L}}}

\def\gR{{\mathcal{R}}}
\def\gS{{\mathcal{S}}}

\def\gX{{\mathcal{X}}}
\def\gY{{\mathcal{Y}}}

\newcommand{\softmax}{\mathrm{softmax}}
\newcommand{\TopK}{\mathrm{TopK}}

\DeclareMathOperator*{\argmax}{arg\,max}

\newcommand{\sigmain}{\sigma_i^n}

\newcommand{\dint}{\mathrm{d}}